\documentclass[10pt,twocolumn,letterpaper]{article}

\usepackage{bm}
\usepackage{booktabs}
\usepackage{pdfpages}
\usepackage{multirow}
\usepackage{subfig}
\newcommand{\mb}[1]{\mathbf{#1}}
\newcommand{\nth}[1]{#1\textsuperscript{th}}
\newcommand{\linetitle}[1]{\noindent\textbf{#1}~~}
\usepackage{iccv}
\usepackage{times}
\usepackage{epsfig}
\usepackage{graphicx}
\usepackage{amsmath}
\usepackage{amssymb}

\usepackage[pagebackref=true,breaklinks=true,letterpaper=true,colorlinks,bookmarks=false]{hyperref}

\iccvfinalcopy % *** Uncomment this line for the final submission

\def\iccvPaperID{2558} % *** Enter the ICCV Paper ID here

\begin{document}

\title{\vspace{-20pt}EigenGAN: Layer-Wise Eigen-Learning for GANs\vspace{-10pt}}

\author{Zhenliang He$^{1,2}$, Meina Kan$^{1,2}$, Shiguang Shan$^{1,2,3}$\\
$^1$ Key Laboratory of Intelligent Information Processing, ICT, CAS\\
$^2$ University of Chinese Academy of Sciences, Beijing, China\\
$^3$ Peng Cheng Laboratory, Shenzhen, China\\
{\tt\small zhenliang.he@vipl.ict.ac.cn, \{kanmeina,sgshan\}@ict.ac.cn}
\vspace{-10pt}
% For a paper whose authors are all at the same institution,
% omit the following lines up until the closing ``}''.
% Additional authors and addresses can be added with ``\and'',
% just like the second author.
% To save space, use either the email address or home page, not both
}

\maketitle
% Remove page # from the first page of camera-ready.
% \ificcvfinal\thispagestyle{empty}\fi

\begin{abstract}
    \vspace{-7pt}
    Recent studies on Generative Adversarial Network (GAN) reveal that different layers of a generative CNN hold different semantics of the synthesized images.
    However, few GAN models have explicit dimensions to control the semantic attributes represented in a specific layer.
    This paper proposes EigenGAN which is able to unsupervisedly mine interpretable and controllable dimensions from different generator layers.
    Specifically, EigenGAN embeds one linear subspace with orthogonal basis into each generator layer.
    Via generative adversarial training to learn a target distribution, these layer-wise subspaces automatically discover a set of ``eigen-dimensions'' at each layer corresponding to a set of semantic attributes or interpretable variations.
    By traversing the coefficient of a specific eigen-dimension, the generator can produce samples with continuous changes corresponding to a specific semantic attribute.
    Taking the human face for example, EigenGAN can discover controllable dimensions for high-level concepts such as pose and gender in the subspace of deep layers, as well as low-level concepts such as hue and color in the subspace of shallow layers.
    Moreover, in the linear case, we theoretically prove that our algorithm derives the principal components as PCA does.
    Codes can be found in \url{https://github.com/LynnHo/EigenGAN-Tensorflow}.
\end{abstract}

\begin{figure}[!t]
    \begin{center}
        \includegraphics[width=1\linewidth]{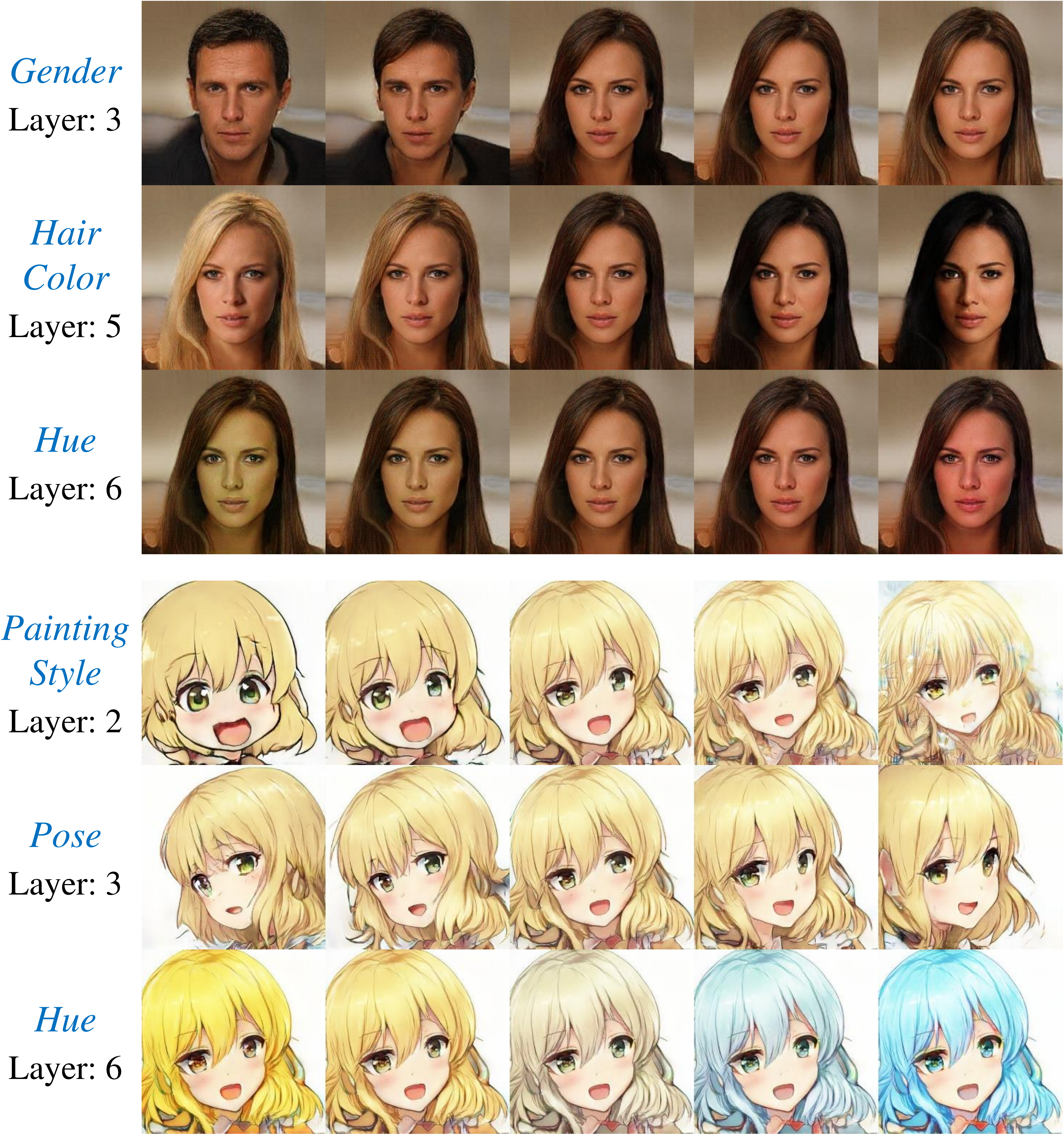}
    \end{center}
    \caption{
        Example of interpretable dimensions learned by EigenGAN.
        The smaller the index, the deeper the layer.
    }\label{fig:first_view}
    \vspace{-10pt}
\end{figure}

\vspace{-10pt}
\section{Introduction}
\vspace{-3pt}

% Generative adversarial network (GAN)~\cite{goodfellow2014generative} and its variants~\cite{mao2017least,gulrajani2017improved,brock2018large,karras2019style} achieve great success in high fidelity image synthesis.
%
Strong evidences~\cite{zeiler2014visualizing,zhou2015object,bau2017network} show that different layers of a discriminative CNN capture different semantic concepts in terms of abstraction level, e.g., shallower layers detect color and texture while deeper layers focus more on objects and parts.
Accordingly, we can expect that a generative CNN also has similar property, which is confirmed by the recent studies of generative adversarial network (GAN)~\cite{karras2019style,yang2021semantic,bau2019gan}.
StyleGAN~\cite{karras2019style} shows that deeper generator layers control higher-level attributes such as pose and glasses while shallower layers control lower-level features such as color and edge.
Yang \etal~\cite{yang2021semantic} found similar phenomenon in scene synthesis, showing that deep layers tend to determine the spatial layout while shallow layers determine the color scheme.
Similar conclusion is also made by Bau \etal~\cite{bau2019gan}.
% Similar conclusion is also made by Bau \etal~\cite{bau2019gan} in the dissection analysis of GAN features at different layers.
%
All these evidences reveal a property that \textit{different generator layers hold different semantics of the synthesized images in terms of abstraction level}.

According to this property, one can identify semantic attributes from different layers of a well-trained generator by performing \textit{post-processing} algorithms~\cite{bau2019gan,harkonen2020ganspace,shen2021closed,yang2021semantic}, and then can manipulate these attributes on the synthesized images.
For example, Bau \etal~\cite{bau2019gan} identify the causal units for a specific concept (such as ``tree'') by dissection and intervention on each generator layer.
Turning on or off the causal units causes the concept to appear or disappear on the synthesized image.
However, these post-processing methods can only be applied to a well-trained and fixed generator.
As for the generator itself, it still operates as a black box and lacks explicit dimensions to directly control the semantic attributes represented in different layers.
In other words, we do not know what attributes are represented in different generator layers or how to manipulate them, unless we deeply inspect each layer by these post-processing methods.

Under above discussion, this paper starts with a question: \textit{can a generator itself automatically/unsupervisedly learn explicit dimensions that control the semantic attributes represented in different layers?}
To this end, we propose to embed one linear subspace model with orthogonal basis into each generator layer, named as EigenGAN.
First, via generative adversarial training, the generator tries to capture the principal variations of the data distribution, and these principal variations are separately represented in different layers in terms of their abstraction level.
Second, with the help of the subspace model, the principal variations of a specific layer are further orthogonally separated into different basis vectors.
Finally, each basis vector discovers an ``eigen-dimension'' that controls an attribute or interpretable variation corresponding to the semantics of its layer.
For example, as shown at the top of Fig.~\ref{fig:first_view}, an eigen-dimension of the subspace embedded in a deep layer controls gender, while another of the subspace embedded in the shallowest layer controls the hue of the image.
Furthermore, in the linear case, i.e., one layer model, we theoretically prove that our EigenGAN is able to discover the principal components as PCA~\cite{jolliffe1986principal} does, which gives us a strong insight and reason to embed the subspace models into different generator layers.
Besides, we also provide a manifold perspective showing that our EigenGAN decomposes the data generation modeling into layer-wise dimension expanding steps.

\section{Related Works}

\subsection{Interpretability Learning for GANs}

The first attempt to learn interpretable representations for GAN generators is InfoGAN~\cite{chen2016infogan} which employs mutual information maximization (MIM) between the latent variable and synthesized samples.
Including InfoGAN, MIM based methods~\cite{chen2016infogan,kaneko2017generative,kaneko2018generative,jeon2021ibgan,lee2020high,lin2020infogan,liu2020oogan} can automatically discover interpretable dimensions which respectively control different semantic attributes such as pose, glasses and emotion of human face.
However, the learning of these interpretable dimensions is mainly driven by the MIM objective, and there is no direct link from these dimensions to the semantics of any specific generator layer.
Ramesh et al.~\cite{ramesh2018spectral} found that the principal right-singular subspace of the generator Jacobian shows local disentanglement property, then they apply a spectral regularization to align the singular vectors with straight coordinates, and finally obtain globally interpretable representations.
However, this work also does not investigate the correspondence between these interpretable representations and the semantics of different generator layers.
Different from these methods, the interpretability of our EigenGAN comes from the special design of layer-wise subspace embedding, rather than imposing any objective or regularization.
Moreover, our EigenGAN establishes an explicit connection between the interpretable dimensions and the semantics of a specific layer by directly embedding a subspace model into that layer.

The above methods try to learn a GAN generator with explicit interpretable representations; in contrast, another class of methods, post-processing methods, try to reveal the interpretable factors from a well-trained GAN generator~\cite{goetschalckx2019ganalyze,bau2019gan,shen2020interfacegan,yang2021semantic,plumerault2019controlling,harkonen2020ganspace,voynov2020unsupervised,shen2021closed}.
\cite{goetschalckx2019ganalyze,bau2019gan,shen2020interfacegan,yang2021semantic} adopt pre-trained semantic predictors to identify the corresponding semantic factors in the GAN latent space, e.g., Yang \etal~\cite{yang2021semantic} use layout estimator, scene category recognizer, and attribute classifier to find out the decision boundaries for these concepts in the latent space.
Without introducing external supervision, several methods search interpretable factors in self-supervised~\cite{plumerault2019controlling} or unsupervised~\cite{harkonen2020ganspace,shen2021closed} manners.
Plumerault \etal~\cite{plumerault2019controlling} utilize simple image transforms (e.g., translation and zoom) to search the axes for these transforms in the latent space.
Harkonen \etal~\cite{harkonen2020ganspace} apply PCA to the feature space of the early layers, and the resulting principal components represent interpretable variations.
Shen and Zhou~\cite{shen2021closed} show that the weight matrix of the very first fully-connected layer of a generator determines a set of critical latent directions which dominate the image synthesis, and the moving along these directions controls a set of semantic attributes.
Among these methods, \cite{bau2019gan,shen2020interfacegan,yang2021semantic,harkonen2020ganspace,shen2021closed} carefully investigate the semantics represented in different generator layers.
However, these post-processing methods must first learn and fix a GAN generator then learn interpretable dimensions under separated objectives (two steps).
On the contrary, our EigenGAN learns the interpretable dimensions for each generator layer along with the GAN training in an end-to-end manner (one step).
Therefore, our method should have a better optimum because the learning of generator and the learning of interpretable dimensions can interact with each other.

\begin{figure*}[!t]
    \begin{center}
        \includegraphics[width=1\linewidth]{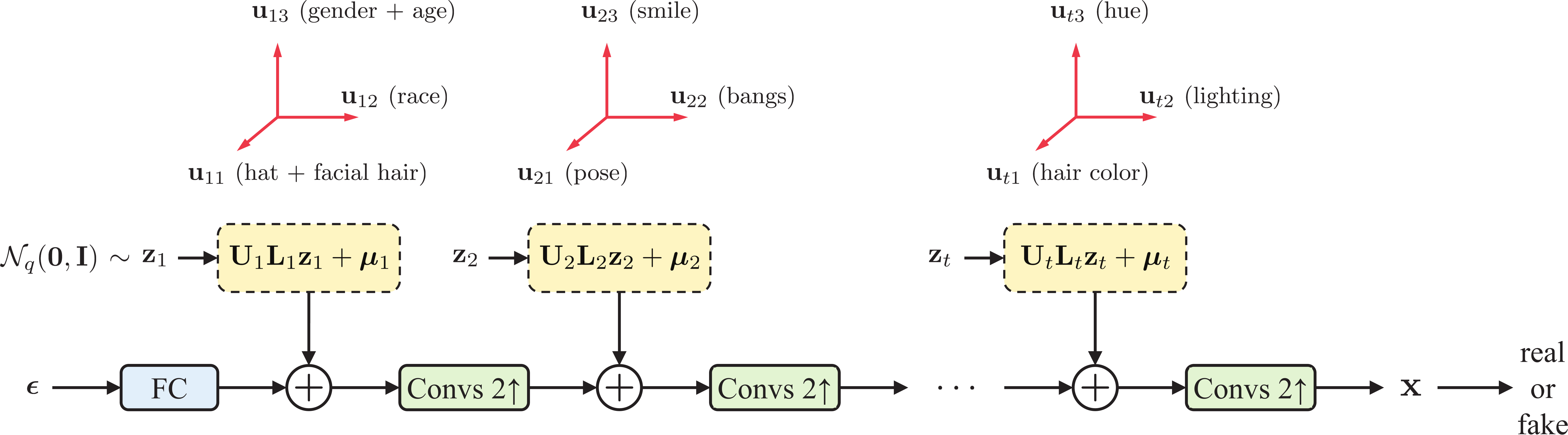}
    \end{center}
    \caption{
        Overview of the proposed EigenGAN.
        The main stream of the model is a chain of 2-stride transposed convolutional blocks which gradually enlarges the resolution of the feature maps and finally outputs a synthesized sample.
        In the \nth{$i$} layer, we embed a linear subspace with orthonormal basis $\mb{U}_i=\left[\mb{u}_{i1}, \ldots, \mb{u}_{iq}\right]$, and each basis vector $\mb{u}_{ij}$ is intended to \textit{\textbf{unsupervisedly}} discover an ``eigen-dimension'' which holds an interpretable variation of the synthesized samples such as race, pose, and lighting for human face.
    }\label{fig:schema}
\end{figure*}

\subsection{Generative Adversarial Networks}

Generative adversarial network (GAN)~\cite{goodfellow2014generative} is a sort of generative model which can synthesize data from noise.
The learning process of GAN is the competition between a generator and a discriminator.
Specifically, the discriminator tries to distinguish the synthesized samples from the real ones, while the generator tries to make the synthesized samples as realistic as possible to fool the discriminator.
When the competition reaches Nash equilibrium, the synthesized data distribution is identical to the real data distribution.

GANs show promising performance and properties on data synthesis.
Therefore, plenty of researches on GANs appear, including loss functions~\cite{nowozin2016f,mao2017least,arjovsky2017wasserstein}, regularizations~\cite{roth2017stabilizing,mescheder2018training,miyato2018spectral}, conditional generation~\cite{mirza2014conditional,odena2017conditional,miyato2018cgans}, representation learning~\cite{makhzani2015adversarial,chen2016infogan,donahue2016adversarial}, architecture design~\cite{denton2015deep,brock2018large,karras2019style}, applications~\cite{isola2017image,zhu2017unpaired,zhang2017stackgan}, and etc.
Our EigenGAN can be categorized into representation learning as well as architecture design for GANs.

\section{EigenGAN}

In this section, we first introduce the EigenGAN generator design with layer-wise subspace models in Sec.~\ref{sec:generator}.
Then in Sec.~\ref{sec:discussion}, we make a discussion from the linear case to the general case of EigenGAN and finally provide a manifold perspective.

\subsection{Generator with Layer-Wise Subspaces}\label{sec:generator}

Fig.~\ref{fig:schema} shows our generator architecture. Our target is to learn a $t$-layer generator mapping from a set of latent variables $\left\{\mb{z}_i\in\mathbb{R}^q\,\middle|\,\mb{z}_i\sim\mathcal{N}_q\left(\bm{0}, \mb{I}\right), i=1,\ldots,t\right\}$ to the synthesized image $\mb{x} = G\left(\mb{z}_1, \ldots, \mb{z}_t\right)$, where $\mb{z}_i$ is directly injected into the \nth{$i$} generator layer; $q$ denotes the number of dimensions of each subspace.

In the \nth{$i$} layer, we embed a linear subspace model $S_i=\left(\mb{U}_i,\mb{L}_i,\bm{\mu}_i\right)$ where
\begin{itemize}
\item $\mb{U}_i=\left[\mb{u}_{i1}, \ldots, \mb{u}_{iq}\right]$ is the orthonormal basis of the subspace, and each basis vector $\mb{u}_{ij}\in\mathbb{R}^{H_i\times W_i \times C_i}$ is intended to unsupervisedly discover an ``eigen-dimension'' which holds an interpretable variation of the synthesized samples.

\item $\mb{L}_i=\text{diag}\left(l_{i1}, \ldots, l_{iq}\right)$ is a diagonal matrix with $l_{ij}$ deciding the ``importance'' of the basis vector $\mb{u}_{ij}$. To be specific, high absolute value of $l_{ij}$ means that $\mb{u}_{ij}$ controls major variation of the the \nth{$i$} layer while low absolute value denotes minor variation, which can be also viewed as a kind of dimension selection.

\item $\bm{\mu}_i$ denotes the origin of the subspace.
\end{itemize}
Then, we use the \nth{$i$} latent variable $\mb{z}_i=\left[z_{i1}, \ldots, z_{iq}\right]^\mathrm{T}$ as the coordinates (linear combination coefficients) to sample a point from the subspace $S_i$:
\begin{align}
\bm{\phi}_i & = \mb{U}_i\mb{L}_i\mb{z}_i + \bm{\mu}_i \label{eq:phi}\\
            & = \sum_{j=1}^q z_{ij}l_{ij}\mb{u}_{ij} + \bm{\mu}_i.
\end{align}
This sample point $\bm{\phi}_i$ will be added to the network feature of the \nth{$i$} layer as stated next.

Let $\mb{h}_i\in\mathbb{R}^{H_i\times W_i \times C_i}$ denote the feature maps of the \nth{$i$} layer and $\mb{x}=\mb{h}_{t+1}$ denote the final synthesized image, then the forward relation between the adjacent layers is
\begin{align}
\mb{h}_{i+1} = \text{Conv2x}\left(\mb{h}_i + f\left(\bm{\phi}_i\right)\right),~~~~~i = 1, \ldots, t, \label{eq:h}
\end{align}
where ``Conv2x'' denotes transposed convolutions that double the resolution of the feature maps, and $f$ can be identity function or a simple transform (1x1 convolution in practice).
As can be seen from Eq.~(\ref{eq:h}), the sample point $\bm{\phi}_i$ from the subspace $S_i$ directly interacts with the network feature $\mb{h}_i$ of the \nth{$i$} layer.
Therefore, the subspace $S_i$ directly determines the variations of the \nth{$i$} layer, more concretely, $q$ coordinates $\mb{z}_i=\left[z_{i1}, \ldots, z_{iq}\right]^\mathrm{T}$ respectively control $q$ different variations.

Besides, we also inject a noise input $\bm{\epsilon}\sim\mathcal{N}\left(\bm{0}, \mb{I}\right)$ to the bottom of the generator intended to capture the rest variations missed by the subspaces, as follows,
\begin{align}
\mb{h}_{1} = \text{FC}\left(\bm{\epsilon}\right),
\end{align}
where ``FC'' denotes the fully-connected layer.

\begin{figure*}[!t]
    \begin{center}
        \includegraphics[width=1\linewidth]{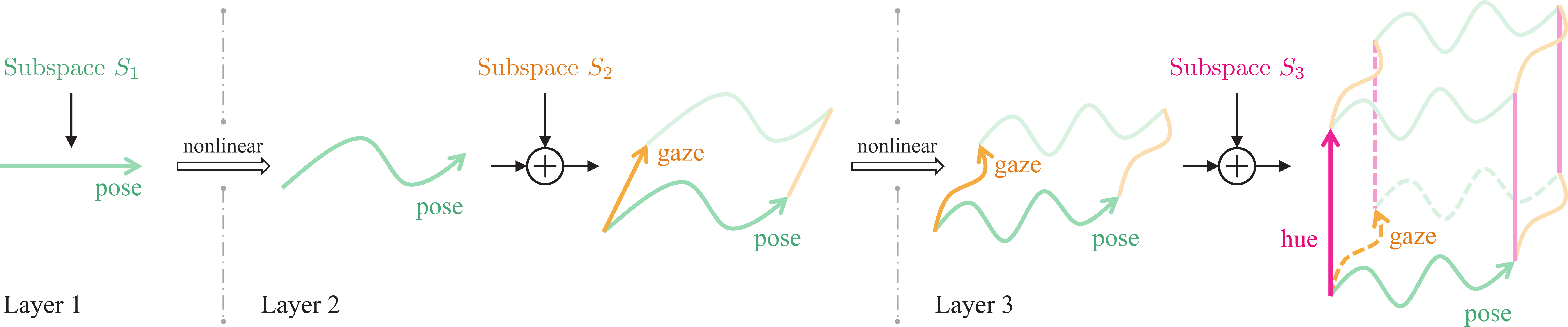}
    \end{center}
    \caption{
        Manifold perspective of EigenGAN.
        At each layer, a linear subspace is added to the feature manifold, expanding the manifold with ``straight'' directions along which the variation of some semantic attributes are linear.
        At the end of each layer, nonlinear mappings ``bend'' these straight directions, yet another subspace at the next layer will continue to add new straight directions.
        Here, we only show one semantic direction of each subspace just for simplicity, generally, each subspace contains multiple orthogonal directions.
    }\label{fig:manifold}
\end{figure*}

The bases $\{\mb{U}_i\}_{i=1}^t$, the importance matrices $\left\{\mb{L}_i\right\}_{i=1}^t$, the origins $\left\{\bm{\mu}_i\right\}_{i=1}^t$, and the convolution kernels are all learnable parameters and the learning can be driven by various adversarial losses~\cite{goodfellow2014generative,mao2017least,arjovsky2017wasserstein,miyato2018spectral}.
In this paper, hinge loss~\cite{miyato2018spectral} is used for the adversarial training.
Besides, the orthogonality of $\mb{U}_i$ is achieved by the regularization of $\|\mb{U}_i^\mathrm{T}\mb{U}_i-\mb{I}\|_F^2$.
After training, each latent dimension $z_{ij}$ can explicitly control an interpretable variation corresponding to the semantic of its layer.

\subsection{Discussion}\label{sec:discussion}

\linetitle{Linear Case} To better understand how our model works, we first discuss the linear case of our EigenGAN.
Adapted from Eq.~(\ref{eq:phi}), the linear model is formulated as below,
\begin{align}
\mb{x} = \mb{U}\mb{L}\mb{z} + \bm{\mu} + \sigma\bm{\epsilon}. \label{eq:linear}
\end{align}
This equation relates a $d$-dimension observation vector $\mb{x}$ to a corresponding $q$-dimension ($q<d$) latent variables $\mb{z}\sim\mathcal{N}_q\left(\bm{0}, \mb{I}\right)$ by an affine transform $\mb{U}\mb{L}$ and a translation $\bm{\mu}$.
Besides, a noise vector $\bm{\epsilon}\sim\mathcal{N}_d\left(\bm{0}, \mb{I}\right)$ is introduce to compensate the missing energy.
We also constrain $\mb{U}$ with orthonormal columns and $\mb{L}$ as a diagonal matrix like the general case in Sec.~\ref{sec:generator}.
This formulation can also be regarded as a constrained case of Probabilistic PCA~\cite{tipping1999probabilistic}.

To estimate $\mb{U}$, $\mb{L}$, $\bm{\mu}$ and $\sigma$ in Eq.~(\ref{eq:linear}) with $n$ observations $\left\{\mb{x}_i\right\}_{i=1}^n$, an analytical solution is maximum likelihood estimation (MLE).
Please refer to the appendix for detailed derivation of the MLE results.
One important result is that the columns of $\mb{U}^{\text{ML}} = \left[\mb{u}^{\text{ML}}_{1}, \ldots, \mb{u}^{\text{ML}}_{q}\right]$ are the eigenvectors of data covariance corresponding to the $q$ largest eigenvalues, which is exactly the same as the result of PCA~\cite{jolliffe1986principal}.
That is to say, \textit{the linear EigenGAN is able to discover the principal dimensions}, which gives us a strong insight and motivation to embed such a linear model (Eq.~(\ref{eq:linear})) hierarchically into different generator layers as stated in Sec.~\ref{sec:generator}.

\medskip\linetitle{EigenGAN (General Case)} With the insight of the linear case, we suppose that the linear subspace model embedded in a specific layer can capture the principal semantic variations of that layer, and these principal variations are orthogonally separated into the basis vectors.
In consequence, each basis vector discovers an ``eigen-dimension'' that controls an attribute or interpretable variation corresponding to the semantics of its layer.

\medskip\linetitle{Manifold Perspective} Fig.~\ref{fig:manifold} shows a manifold perspective of EigenGAN.
From this aspect, the subspace of each layer expands the feature manifold with ``straight'' directions along which the variations of some semantic attributes are linear.
At the end of each layer, nonlinear mappings ``bend'' these straight directions, yet another subspace at the next layer will continue to add new straight directions.
In a word, EigenGAN \textit{decomposes the data generation modeling into hierarchical dimension expanding steps}, i.e., expanding the feature manifold with linear semantic dimensions layer-by-layer.

\begin{figure*}[!t]
    \begin{center}
        \includegraphics[width=1\linewidth]{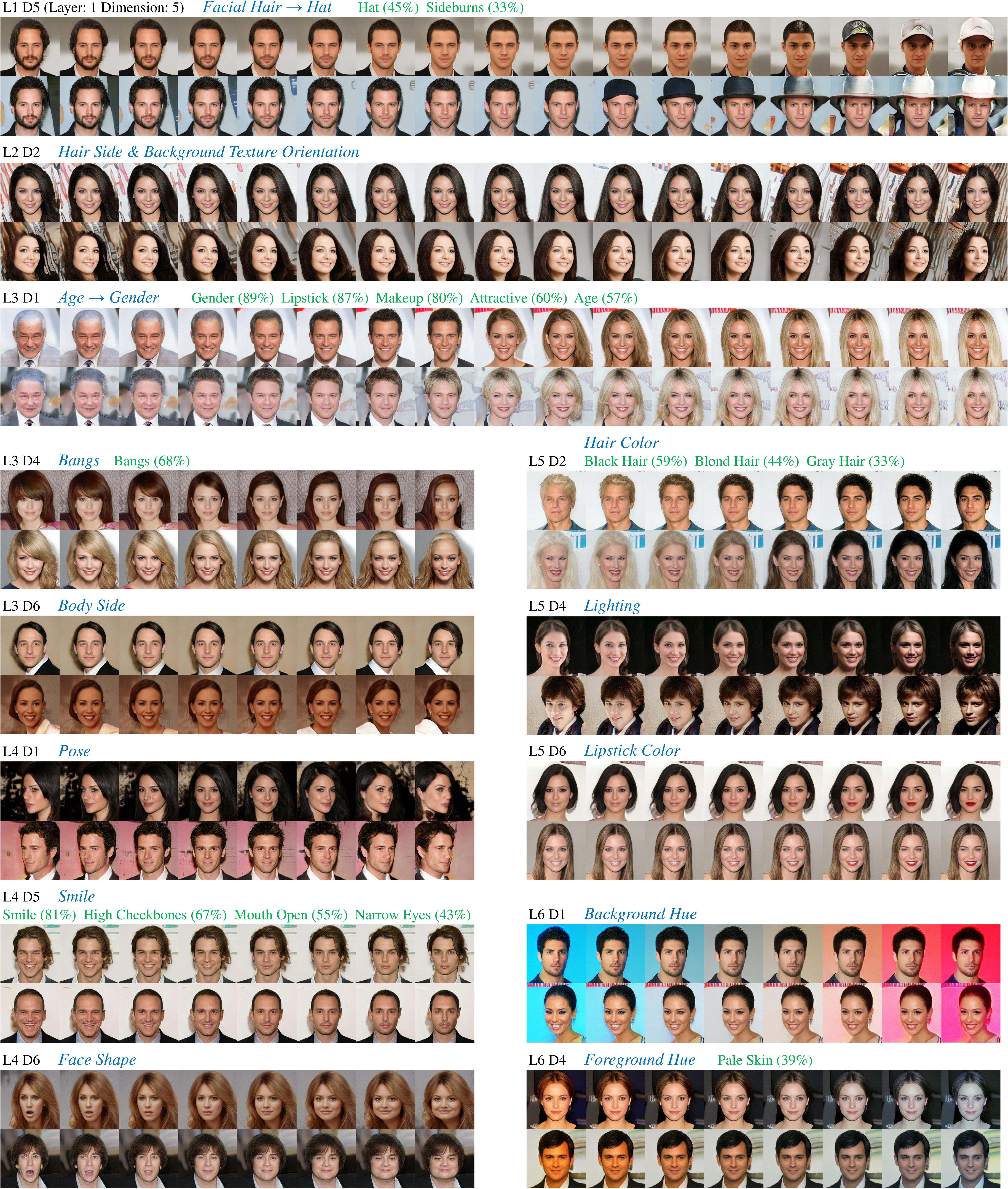}
    \end{center}
    \vspace{-1pt}
    \caption{
        Discovered semantic attributes at different layers for CelebA dataset~\cite{liu2015deep}.
        Traversing the coordinate value in $\left[-4.5\sigma, 4.5\sigma\right]$, each dimension controls an attribute, colored in blue.
        The attributes colored in green are the most correlated CelebA attributes, and the bracket value is the entropy coefficient: what fraction of the information of the CelebA attribute is contained in the corresponding dimension.
        ``Li Dj'' means the \nth{j} dimension of the \nth{i} layer.
        We only show the most meaningful dimensions, please refer to the appendix for all dimensions.
    }\label{fig:dimensions}
\end{figure*}

\begin{figure*}[!t]
    \begin{center}
        \begin{minipage}{0.477\linewidth}
            \begin{center}
                \includegraphics[width=1\linewidth]{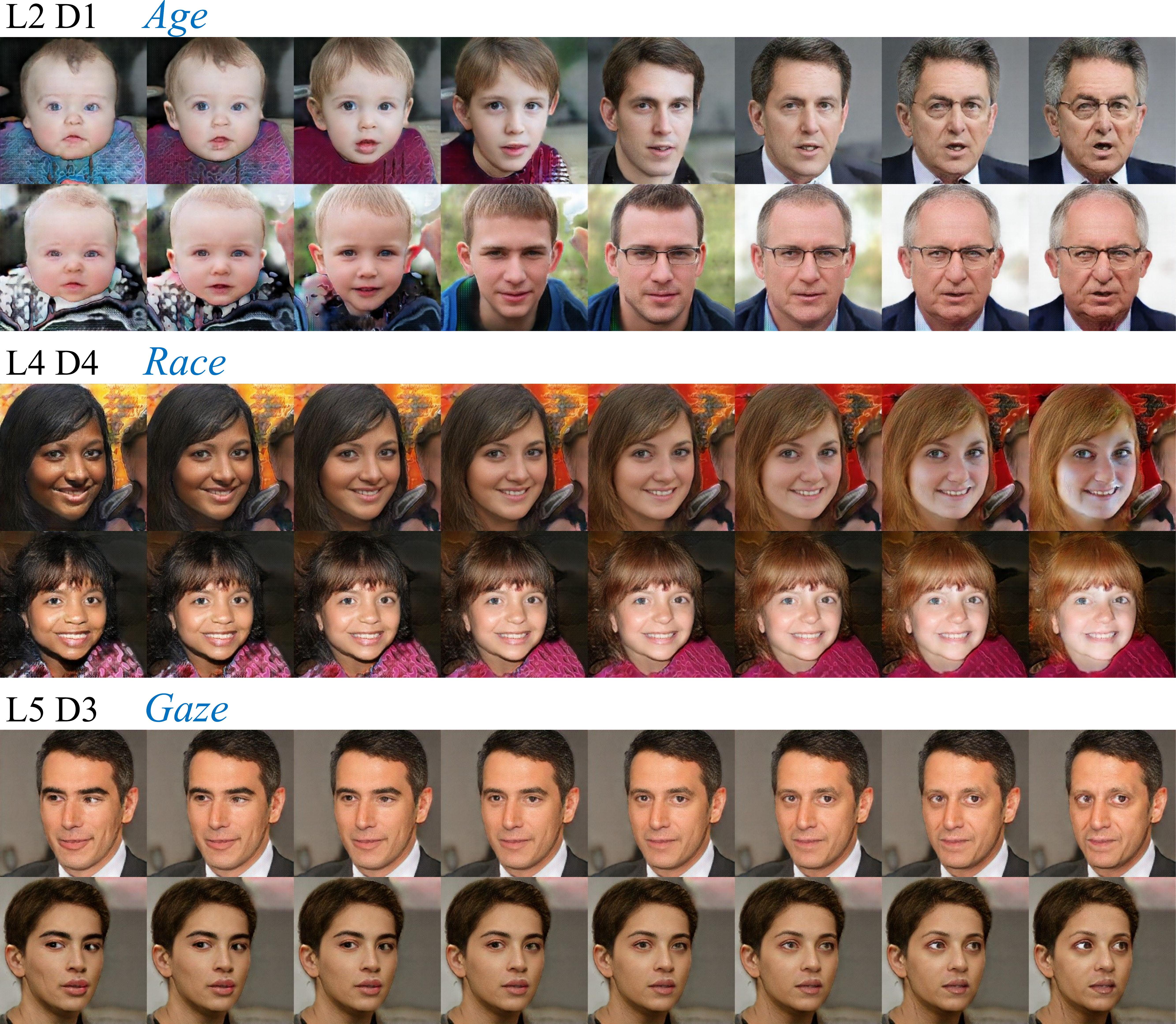}
            \end{center}
        \end{minipage}
        \hfill
        \begin{minipage}{0.477\linewidth}
            \begin{center}
                \includegraphics[width=1\linewidth]{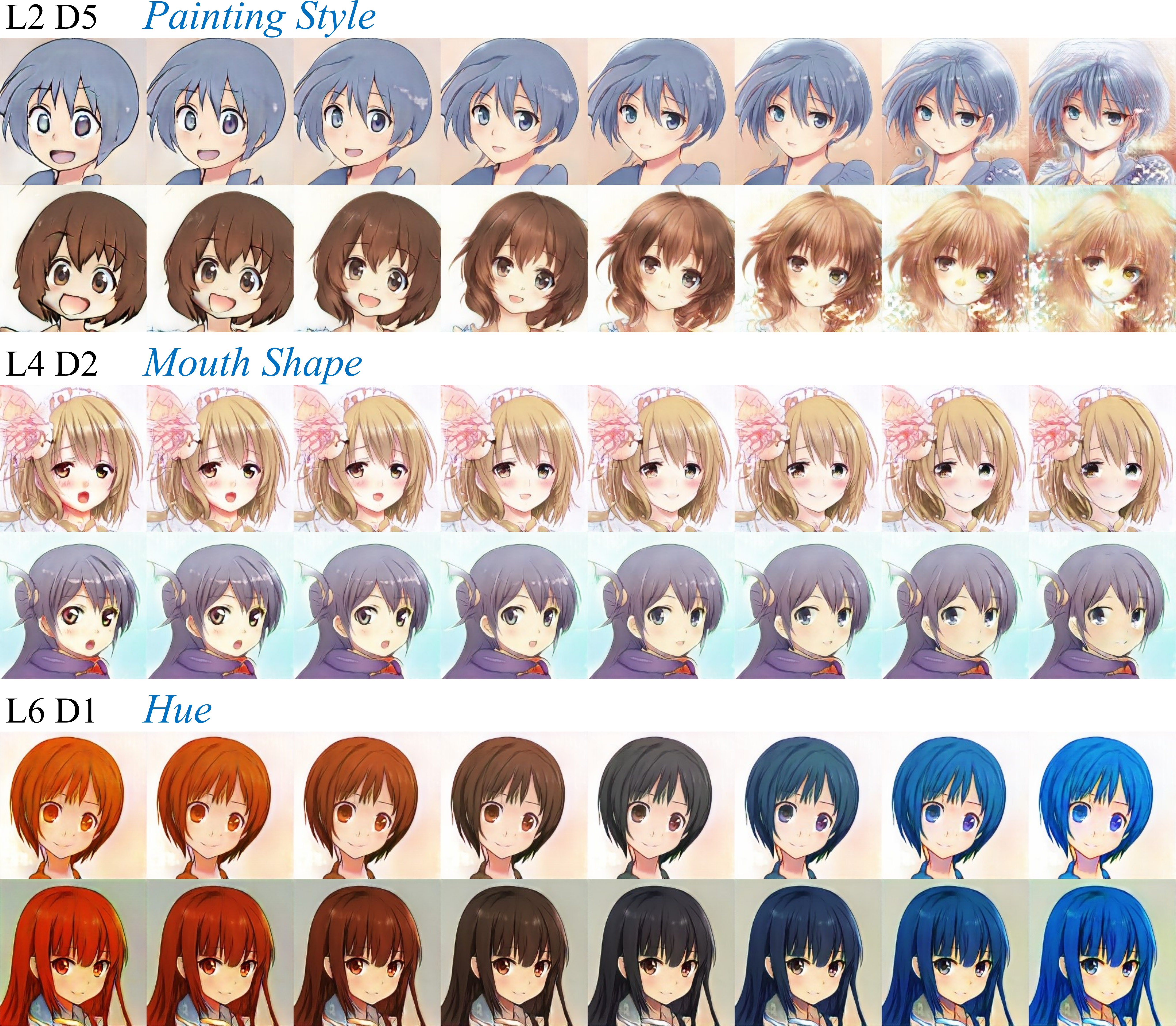}
            \end{center}
        \end{minipage}
    \end{center}
    \caption{
         Interpretable dimensions of FFHQ dataset~\cite{karras2019style} and anime dataset~\cite{danbooru2019Portraits}.
    }\label{fig:dimensions_ffhq_anime}
\end{figure*}

\section{Experiments}

\linetitle{Dataset}
We test our method on CelebA~\cite{liu2015deep}, FFHQ~\cite{karras2019style}, and Danbooru2019 Portraits~\cite{danbooru2019Portraits}.
CelebA contains 202,599 celebrity face images with annotations of 40 binary attributes.
FFHQ contains 70,000 high-quality face images and Danbooru2019 Portraits contains 302,652 anime face images.
We use CelebA attributes for the quantitative evaluations and use FFHQ and Danbooru2019 Portraits for more visual results.

\medskip\linetitle{Implementation Details}
We use hinge loss~\cite{miyato2018spectral} and $R_1$ penalty~\cite{mescheder2018training} for the adversarial training.
We adopt Adam solver~\cite{kingma2015adam} for all networks and parameter moving average for the generator.
The generator is designed for $256\times256$ images and contains 6 upsampling convolutional blocks.
A whole block with one upsampling is defined as a ``layer'', and one linear subspace with 6 basis vectors is embedded into each generator layer.
Please refer to the appendix for detailed network architectures.

\vspace{5pt}
\subsection{Discovered Semantic Attributes}~\label{sec:disco_atts}

\vspace{-10pt}
\linetitle{Visual Analysis}
Fig.~\ref{fig:dimensions} shows the semantic attributes learned by the subspace of different layers, where ``Li Dj'' means the \nth{j} dimension of the \nth{i} layer and smaller index of layer means deeper.
As shown, moving along an eigen-dimension (i.e., a basis vector of a subspace), the synthesized images consistently change by an interpretable meaning.
Shallower layers tend to learn lower-level attributes, e.g., L6 and L5 learn color-related attributes such as ``Hue'' in L6 and ``Hair Color'' in L5.
As the layer goes deeper, the generator discovers attributes with higher-level or more complicated concepts.
For example, L4 and L3 learn geometric or structural attributes such as ``Face Shape'' in L4 and ``Body Side'' in L3.
Deep layers tend to learn multiple attributes in one dimension, e.g., L1 D5 learns ``Facial Hair'' on the left axis but ``Hat'' on the right axis.
Besides, entanglement of attributes is likely to happen in deep layer dimensions, e.g., L2 D2 learns to simultaneously change ``Hair Side'' and ``Background Texture Orientation'', because complex attribute composition might mislead the network into believing their whole as one high-level attribute.

In summary, shallow layers learn low-level or simple attributes while deep layers learn high-level or complicated attributes.
Entanglement might happen in some dimensions of deep layers, and this is one of our limitations.
Nonetheless, the entanglement is interpretable, i.e., we can identify what attributes are entangled in a dimension.
Moreover, our method can still discover well disentangled dimensions that are highly consistent with the visual concepts of humans.
Fig.~\ref{fig:dimensions_ffhq_anime} show additional results of FFHQ dataset~\cite{karras2019style} and Danbooru2019 Portraits dataset~\cite{danbooru2019Portraits}.
Please refer to the appendix for more results and more interpretable dimensions.

\begin{figure*}[!t]
    \begin{center}
        \includegraphics[width=1\linewidth]{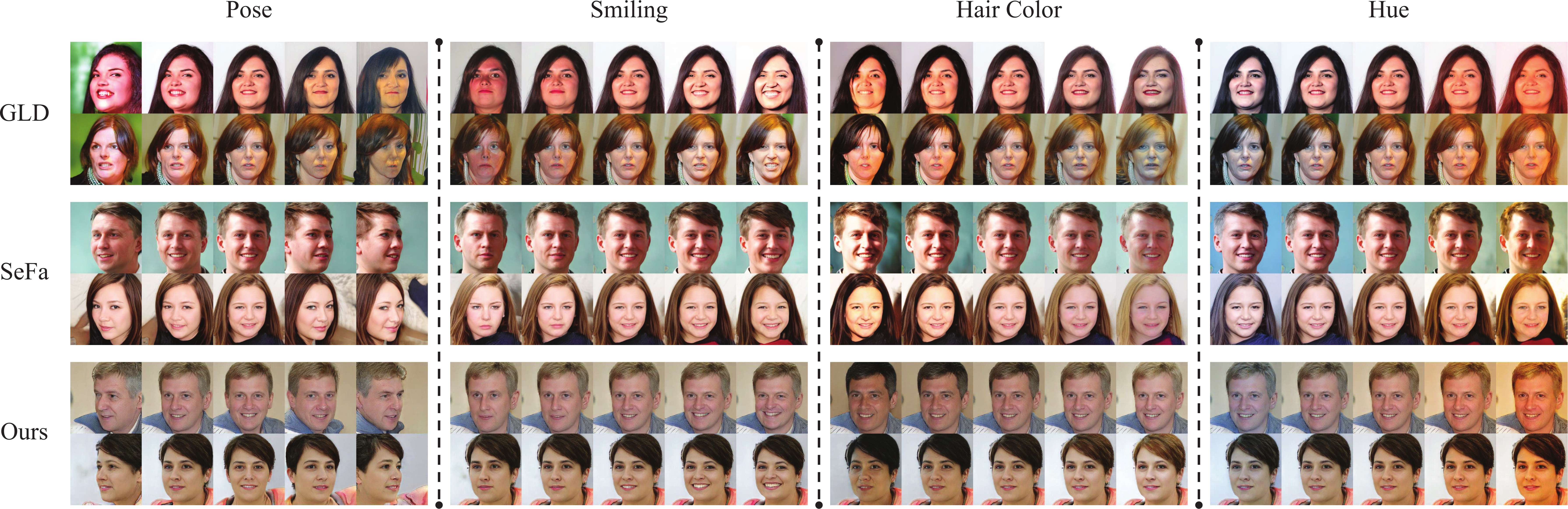}
    \end{center}
    \vspace{-10pt}
    \caption{
        Qualitative comparison among GLD~\cite{voynov2020unsupervised}, SeFa~\cite{shen2021closed}, and our EigenGAN.
    }\label{fig:sefa_comp}
    \vspace{-5pt}
\end{figure*}

\begin{table*}[!t]
    \caption{
        Correlation between the discovered attributes and the CelebA attributes in terms of entropy coefficient.
        Each row denotes a discovered attributes by GLD~\cite{voynov2020unsupervised}, SeFa~\cite{shen2021closed} and our EigenGAN, and each column denotes a CelebA attribute.
    }~\label{tab:sefa_comp}
    \vspace{-15pt}
    \begin{center}
        \includegraphics[width=0.32\linewidth]{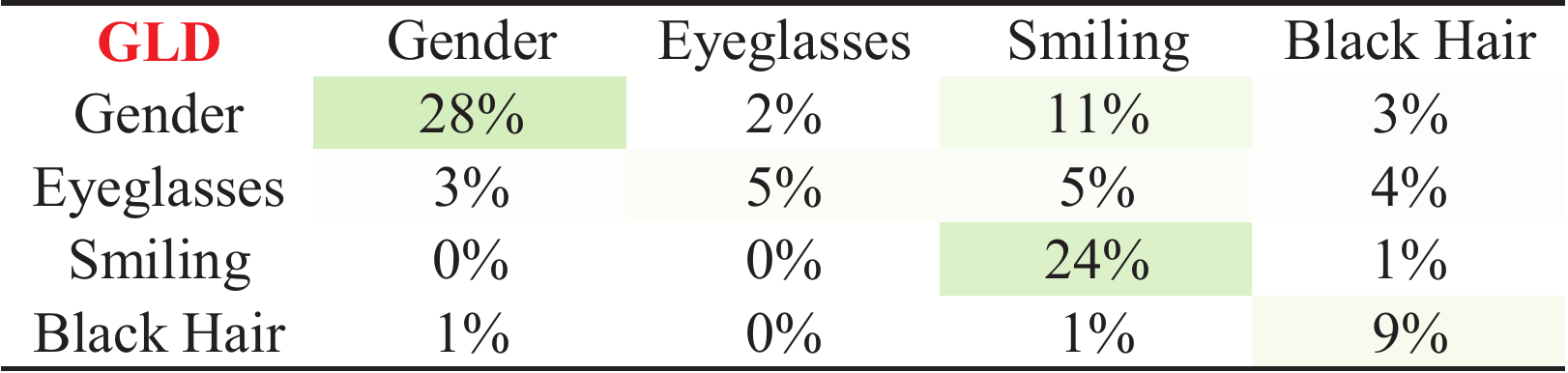}
        \hfill
        \includegraphics[width=0.32\linewidth]{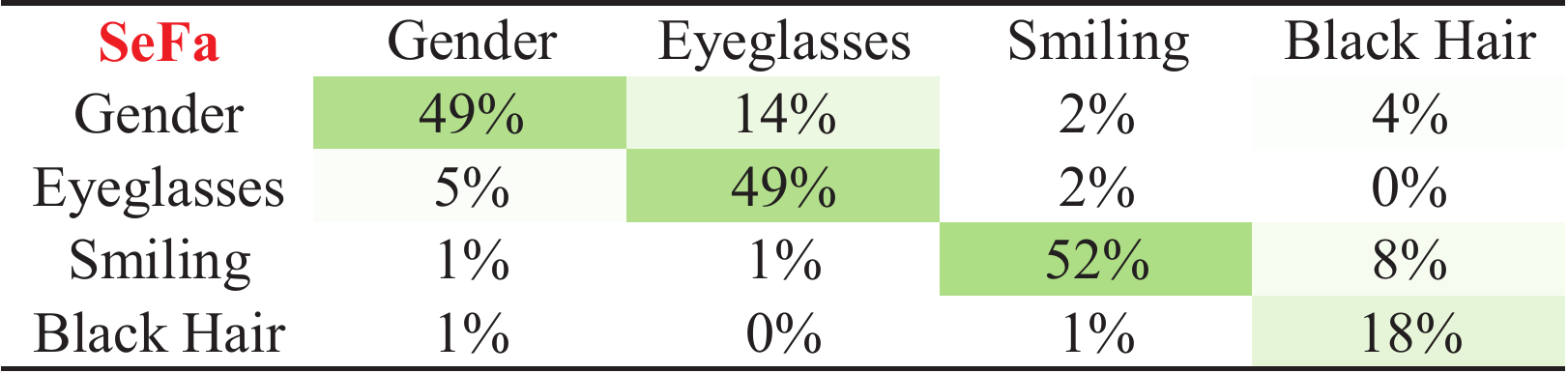}
        \hfill
        \includegraphics[width=0.32\linewidth]{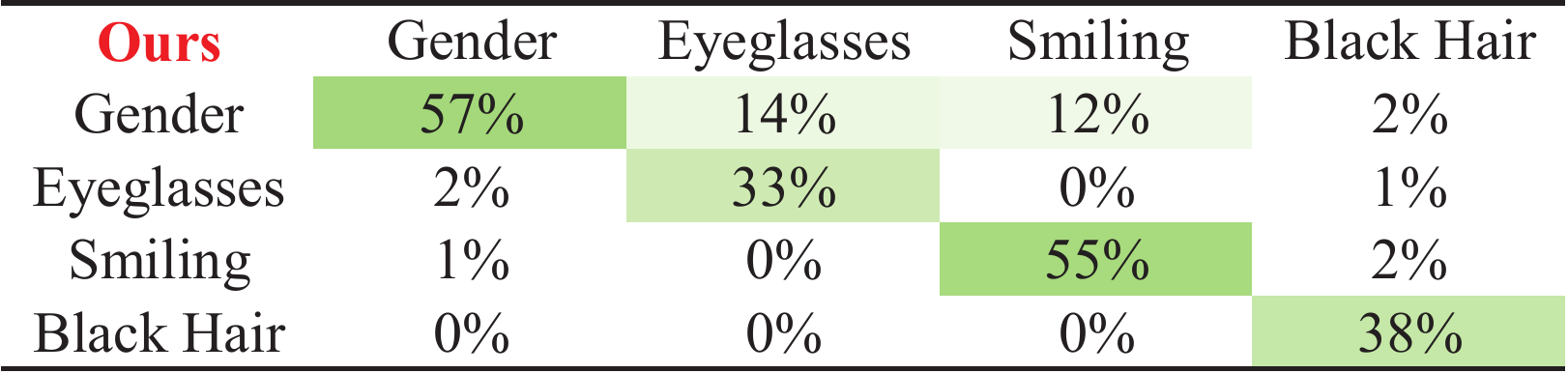}
    \end{center}
    \vspace{-20pt}
\end{table*}

\medskip\linetitle{Identifying Well-Defined Attributes}
In the previous part, we visually identify semantic attributes for each dimension.
In this part, we identify the attributes in a statistical manner, utilizing 40 well-defined binary attributes in CelebA dataset~\cite{liu2015deep}.
Specifically, we investigate the correlation between a dimension $Z$ and a CelebA attribute $Y$ in terms of \textit{entropy coefficient} (normalized mutual information), which represents what fraction of the information of $Y$ is contained in $Z$:
\begin{align}
\mb{U}\left(Y|Z\right)=\frac{\mb{I}(Y;Z)}{\mb{H}(Y)} = \frac{\mb{H}(Y) - \mb{H}(Y|Z)}{\mb{H}(Y)}\in [0,1]
\end{align}
where
\begin{align}
\mb{H}(Y|Z) = & \int_{\!\mathcal{Z}}p_{\scalebox{0.5}{Z}}(z)\big[-p_{\scalebox{0.5}{Y}{\scriptscriptstyle|}\scalebox{0.5}{Z}}(y\!=\!1|z)\ln\left(p_{\scalebox{0.5}{Y}{\scriptscriptstyle|}\scalebox{0.5}{Z}}(y\!=\!1|z)\right) \nonumber\\
              & \!\!\!\!\!\!\!\!\!\!\!\!\!\!\!-\left(1-p_{\scalebox{0.5}{Y}{\scriptscriptstyle|}\scalebox{0.5}{Z}}(y\!=\!1|z)\right)\ln\left(1-p_{\scalebox{0.5}{Y}{\scriptscriptstyle|}\scalebox{0.5}{Z}}(y\!=\!1|z)\right)\big]dz, \label{eq:h_y_z} \\[10pt]
\mb{H}(Y) =   & -p_{\scalebox{0.5}{Y}}(y\!=\!1)\ln\left(p_{\scalebox{0.5}{Y}}(y\!=\!1)\right) \nonumber\\
              & -\left(1-p_{\scalebox{0.5}{Y}}(y\!=\!1)\right)\ln\left(1-p_{\scalebox{0.5}{Y}}(y\!=\!1)\right). \label{eq:h_y}
\end{align}
$p_{\scalebox{0.5}{Y}{\scriptscriptstyle|}\scalebox{0.5}{Z}}(y\!=\!1|z)$ and $p_{\scalebox{0.5}{Y}}(y\!=\!1)$ can be calculated by\footnote{$y$ and $z$ are conditionally independent given $x$, i.e., $p_{\scalebox{0.5}{Y}{\scriptscriptstyle|}\scalebox{0.5}{X,Z}}(y\!=\!1|x,z)=p_{\scalebox{0.5}{Y}{\scriptscriptstyle|}\scalebox{0.5}{X}}(y\!=\!1|x)$.}
\begin{align}
p_{\scalebox{0.5}{Y}{\scriptscriptstyle|}\scalebox{0.5}{Z}}(y\!=\!1|z) = & \int_{\!\mathcal{X}}p_{\scalebox{0.5}{Y}{\scriptscriptstyle|}\scalebox{0.5}{X}}(y\!=\!1|x)p_{\scalebox{0.5}{G}}(x|z)dx, \label{eq:p_y_z}\\
p_{\scalebox{0.5}{Y}}(y\!=\!1) =                                          & \int_{\!\mathcal{Z}}p_{\scalebox{0.5}{Y}{\scriptscriptstyle|}\scalebox{0.5}{Z}}(y\!=\!1|z)p_{\scalebox{0.5}{Z}}(z)dz, \label{eq:p_y}
\end{align}
where $p_{\scalebox{0.5}{G}}(x|z)$ is the generator distribution, and $p_{\scalebox{0.5}{Y}{\scriptscriptstyle|}\scalebox{0.5}{X}}(y\!=\!1|x)$ is the posterior distribution which is approximated by a pre-trained attribute classifier on CelebA dataset.
%`
We set $p_{\scalebox{0.5}{Z}}(z)$ as $\mathcal{U}[-4.5,4.5]$ and discretize it into 100 equal bins for approximation of the integral $\int_{\!\mathcal{Z}}\cdot\,\,p_{\scalebox{0.5}{Z}}(z)dz$ in Eq.~(\ref{eq:h_y_z}) and (\ref{eq:p_y});
and we sample 1000 $x$ from the generator $p_{\scalebox{0.5}{G}}(x|z)$ in each bin of $z$, then approximate the integral $\int_{\!\mathcal{X}}\cdot\,\,p_{\scalebox{0.5}{G}}(x|z)dx$ in Eq.~(\ref{eq:p_y_z}) by averaging over the samples.

For each dimension in Fig.~\ref{fig:dimensions}, the five most correlated CelebA attributes with entropy coefficient larger than 30\% are shown (green text).
As shown, the identified CelebA attributes according to entropy coefficient are highly consistent with our visual perception.
Several dimensions have no correlated CelebA attributes just because the attributes represented by these dimensions are not included in the CelebA, but these dimensions are still interpretable, e.g., L4 D1 learns ``Pose'' which is not a CelebA attribute.
Several dimensions correlate to multiple CelebA attributes mainly because these CelebA attributes are themselves highly correlated, e.g., L4 D5 learns ``Smile'' therefore it has high entropy coefficient for ``Smile'' correlated attributes: ``High Cheekbones'', ``Mouth Open'', and ``Narrow Eyes''.
In conclusion, this experiment statistically verifies that, EigenGAN can indeed discover interpretable dimensions controlling attributes which are highly consistent with human-defined ones (e.g., CelebA attributes).

\medskip\linetitle{Comparison}
In this part, we compare our method to two state-of-the-art post-processing methods GANLatentDiscovery (GLD)~\cite{voynov2020unsupervised} and SeFa~\cite{shen2021closed}.
We use their official models with GLD trained on StyleGAN2-FFHQ-1024 and SeFa trained on StyleGAN-FFHQ-256.
Fig.~\ref{fig:sefa_comp} shows the qualitative comparison.
As can be seen, both SeFa and our EigenGAN can achieve smooth and consistent change of the identified attributes, more natural and realistic than GLD.
However, entanglement to some extent still happens in all three methods, e.g., ``Pose'' dimension also changes lighting in GLD, ``Smiling'' dimension also changes bangs in SeFa, and ``Hair Color'' dimension also changes skin color in EigenGAN.
This is because all of them are unsupervised methods, and it is difficult to precisely decouple all the attributes without any supervision.
Table~\ref{tab:sefa_comp} shows the quantitative comparison of the correlation between the discovered attributes and the CelebA attributes, in terms of entropy coefficient introduced in the previous part.
As can be seen, the discovered attributes by both SeFa and our EigenGAN have high correlation to the corresponding CelebA attributes, demonstrating that both methods can indeed discover meaningful semantic attributes.
Overall, our EigenGAN achieves comparable performance to the state-of-the-art SeFa on the learned attributes and disentanglement, and both methods perform better than GLD.

\begin{figure*}[!t]
    \captionsetup{farskip=0pt}
    \begin{center}
        \subfloat[With the subspace models (EigenGAN), major variations are captured by the layer-wise latent variables.]{\includegraphics[width=0.49\linewidth]{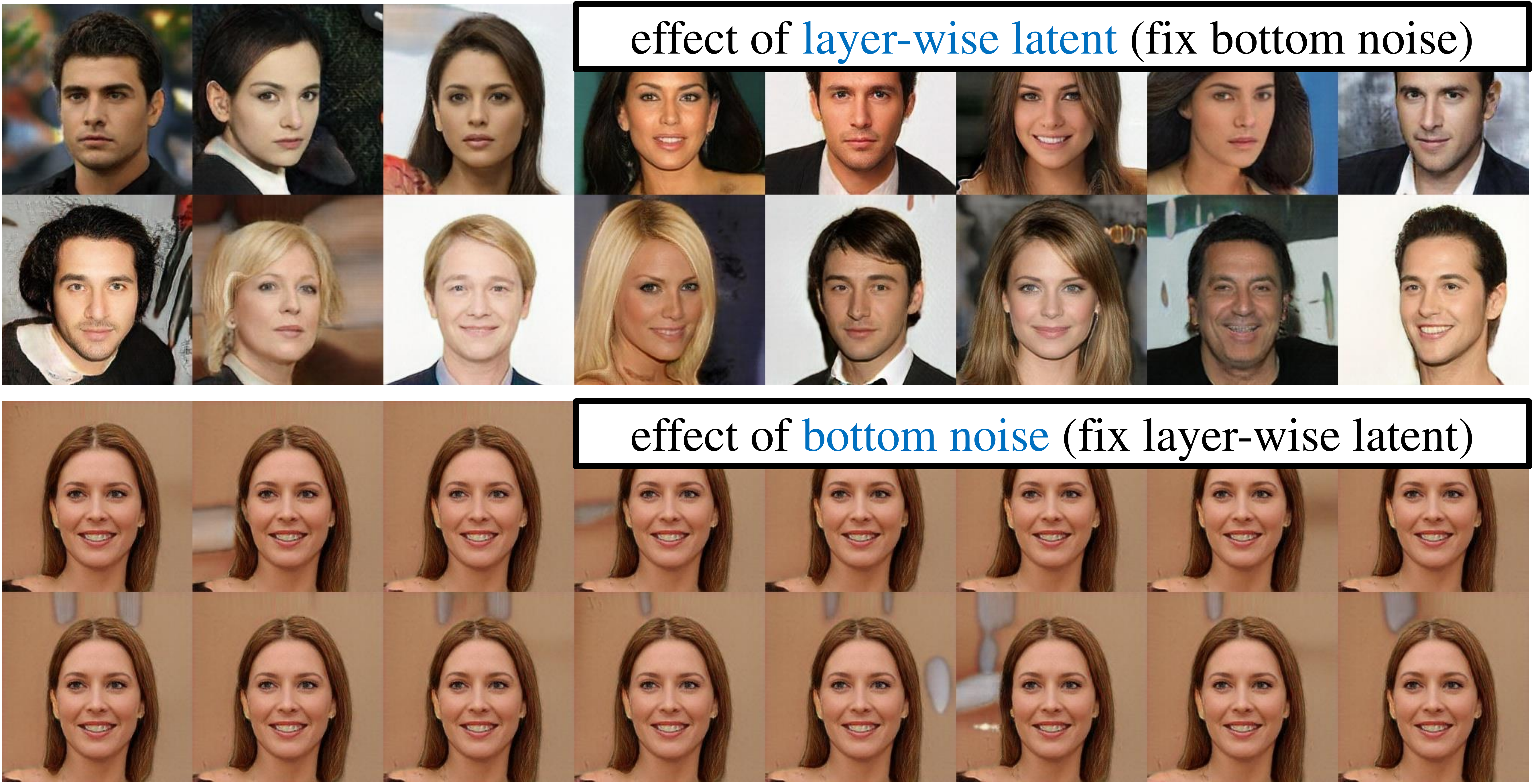}\label{fig:fix_eps_z}}
        \hfill
        \subfloat[Without the subspace models (typical GANs), major variations are captured by the bottom noise.]{\includegraphics[width=0.49\linewidth]{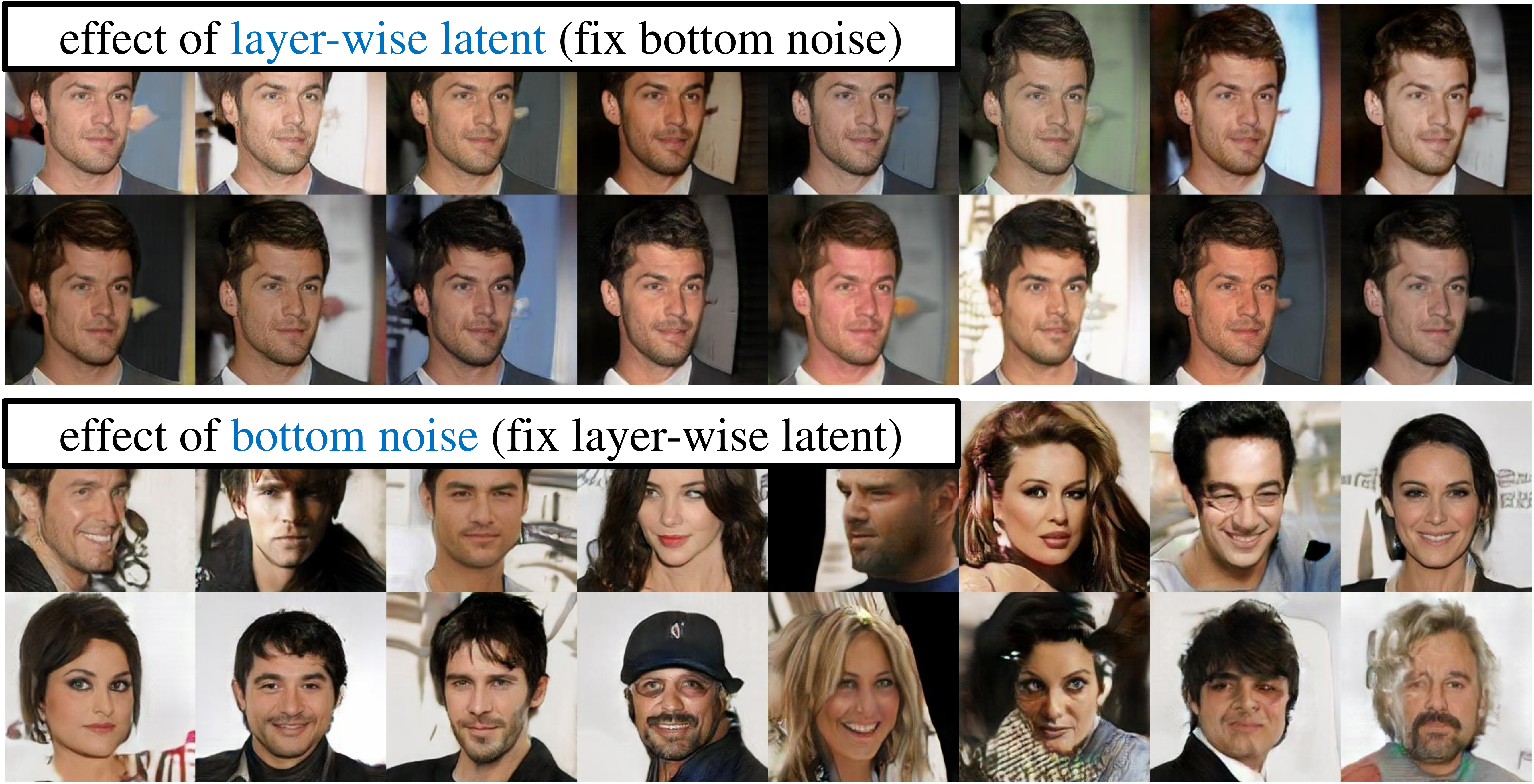}\label{fig:fix_eps_z_direct_add}}
    \end{center}
    \vspace{-5pt}
    \caption{
        Effect of the layer-wise latent variables (top) and the bottom noise (down).
    }
    \vspace{-5pt}
\end{figure*}

\begin{table*}[!t]
    \begin{center}
        \begin{minipage}{0.477\linewidth}
            \caption{
                Basis similarity with PCA. $P=\mathcal{N}_d\left(\mb{0}, \mb{I}\right)$.
            }\label{tab:normal}
            \begin{center}
                \resizebox{1\linewidth}{!}{%
                    \begin{tabular}{lcccccccc}
                        \cmidrule[2\arrayrulewidth]{1-9}
                        \multirow{2}{*}[-2pt]{GAN Loss} & \multicolumn{8}{c}{Data Rank → Subspace Rank}             \\
                        \cmidrule{2-9}
                                                                    & 5→1  & 5→3  & 10→1 & 10→3 & 10→5 & 20→1 & 20→5 & 20→10  \\
                        \cmidrule{1-9}
                        KL-f-GAN~\cite{nowozin2016f}                & 1.00 & 0.98 & 0.99 & 0.90 & 0.93 & 0.97 & 0.78 & 0.79   \\
                        Vanilla GAN~\cite{goodfellow2014generative} & 1.00 & 0.99 & 1.00 & 0.90 & 0.94 & 0.98 & 0.77 & 0.81   \\
                        WGAN~\cite{gulrajani2017improved}           & 0.99 & 0.98 & 1.00 & 0.89 & 0.92 & 0.99 & 0.76 & 0.83   \\
                        LSGAN~\cite{mao2017least}                   & 0.99 & 0.99 & 1.00 & 0.89 & 0.92 & 0.99 & 0.76 & 0.80   \\
                        HingeGAN~\cite{miyato2018spectral}          & 0.99 & 0.99 & 1.00 & 0.92 & 0.93 & 0.96 & 0.77 & 0.81   \\
                        \cmidrule[2\arrayrulewidth]{1-9}
                    \end{tabular}
                }
            \end{center}
        \end{minipage}
        \hfill
        \begin{minipage}{0.477\linewidth}
            \caption{
                Basis similarity with PCA. $P=\mathcal{U}_d(0, 1)$.
            }\label{tab:uniform}
            \begin{center}
                \resizebox{1\linewidth}{!}{%
                    \begin{tabular}{lcccccccc}
                        \cmidrule[2\arrayrulewidth]{1-9}
                        \multirow{2}{*}[-2pt]{GAN Loss} & \multicolumn{8}{c}{Data Rank → Subspace Rank}             \\
                        \cmidrule{2-9}
                                                                    & 5→1  & 5→3  & 10→1 & 10→3 & 10→5 & 20→1 & 20→5 & 20→10  \\
                        \cmidrule{1-9}
                        KL-f-GAN~\cite{nowozin2016f}                & 0.96 & 0.98 & 0.97 & 0.89 & 0.93 & 0.89 & 0.72 & 0.82   \\
                        Vanilla GAN~\cite{goodfellow2014generative} & 0.97 & 0.97 & 0.97 & 0.92 & 0.92 & 0.92 & 0.76 & 0.84   \\
                        WGAN~\cite{gulrajani2017improved}           & 0.98 & 0.97 & 0.98 & 0.93 & 0.94 & 0.98 & 0.77 & 0.84   \\
                        LSGAN~\cite{mao2017least}                   & 0.97 & 0.97 & 0.96 & 0.89 & 0.95 & 0.91 & 0.74 & 0.82   \\
                        HingeGAN~\cite{miyato2018spectral}          & 0.97 & 0.98 & 0.97 & 0.87 & 0.94 & 0.92 & 0.75 & 0.82   \\
                        \cmidrule[2\arrayrulewidth]{1-9}
                    \end{tabular}
                }
            \end{center}
        \end{minipage}
    \end{center}
    \vspace{-20pt}
\end{table*}

\subsection{Model Analysis}

\linetitle{Effect of the Latent Variables}
EigenGAN contains two kinds of latent variables: 1) layer-wise latent variables $\left\{\mb{z}_i\right\}_{i=1}^t$, which are used as the subspace coordinates; 2) bottom noise $\bm{\epsilon}$ to compensate the missing variations.
In Fig.~\ref{fig:fix_eps_z}, we respectively fix one of them and randomly sample another to generate images.
As can be seen, the layer-wise latent variables $\left\{\mb{z}_i\right\}_{i=1}^t$ dominate the major variations while the bottom noise $\bm{\epsilon}$ captures subtle changes.
That is to say, EigenGAN tends to put major variations into the layer-wise latent variables rather than the bottom noise used in typical GANs, but the bottom noise can still capture some subtle variations missed by the subspace models.

\medskip\linetitle{Effect of the Subspace Model}
We remove all the layer-wise subspace models to investigate their effect, instead, we directly add the layer-wise latent variables to the network features.
As shown in Fig.~\ref{fig:fix_eps_z_direct_add}, without the subspace models, the layer-wise latent variables can only capture minor variations, which is completely opposite to the original setting in Fig.~\ref{fig:fix_eps_z}.
In conclusion, the subspace model is the key point to enable the generator to put major variations into the layer-wise variables, therefore can further let the layer-wise variables capture different semantics of different layers.

\medskip\linetitle{Linear Case Study}
Sec.~\ref{sec:discussion} theoretically proves that the linear case of EigenGAN can discover the principal components under maximum likelihood estimation (MLE).
In this part, we validate this statement by applying adversarial training on the linear EigenGAN (we do not directly use MLE since we train the general EigenGAN with adversarial loss rather than MLE objective, and we keep this consistency between the linear and the general case).
Specifically, we use the linear EigenGAN to learn a low-rank subspace model for toy datasets, then compare the basis vectors learned by our model and learned by PCA in terms of cosine similarity.
The toy datasets are generated as follows,
\begin{align}
\mathcal{D}_{\mb{A},\mb{b},P} = \left\{\mb{y}_i = \mb{A}\mb{x}_i+\mb{b}~\middle|~ \mb{x}_i\sim P\right\}
\end{align}
where $\mb{A}$ is a random transform matrix, $\mb{b}$ is a random translation vector, and $P$ is a distribution selected from $\mathcal{N}_d\left(\mb{0}, \mb{I}\right)$ or $\mathcal{U}_d(0,1)$.
We test typical adversarial losses including Vanilla GAN~\cite{goodfellow2014generative}, LSGAN~\cite{mao2017least}, WGAN~\cite{gulrajani2017improved}, HingeGAN~\cite{miyato2018spectral}, and f-GAN~\cite{nowozin2016f} with KL divergence (KL-f-GAN).
Note that the objective of KL-f-GAN is theoretically equivalent to MLE, thus we are actually also testing MLE in the adversarial training manner.

Table~\ref{tab:normal} and Table~\ref{tab:uniform} report the average similarity between EigenGAN basis vectors and PCA basis vectors, where each result is the average over 100 random toy datasets.
As can be seen, when the data rank is no more than 10, EigenGAN basis is highly similar to PCA basis with cosine similarity of about 0.9-1.0.
When the data rank increases to 20, there are two situations: 1) if we only search the most principal one basis vector (20→1), the vectors found by linear EigenGAN and by PCA are still very close; 2) but if we want to find 5 or more basis vectors, the average similarity decreases to 0.7-0.8.
We suppose the reason is that higher dimension data leads to the curse of dimensionality and further results in learning instability.
Besides, various GAN losses have very consistent results, which shows the potential of generalizability of our theoretical results in Sec.~\ref{sec:discussion} from  KL divergence (MLE) to more general statistical distance such JS divergence and Wasserstein distance.
In conclusion, we experimentally verify the theoretical statement that the linear EigenGAN can indeed discover principal components.

\section{Limitations and Future Works}

Discovered semantic attributes are not always the same at different training times in two cases: 1) E.g., sometimes the gender and pose are learned as separated dimensions but sometimes are entangled in one dimension at a deeper layer. This is because, without supervision, some complex attribute compositions might mislead the model into believing their whole as one higher-level attribute.
2) Sometimes the model can discover a specific attribute but sometimes cannot, such as eyeglasses, mainly because these attributes appear less frequently in the dataset.
Future works will study the layer-wise eigen-learning with better disentanglement techniques and more powerful GAN architectures.

\medskip\linetitle{Acknowledgement}
This work is partially supported the National Key Research and Development Program of China (No. 2017YFA0700800) and the Natural Science Foundation of China (No. 61772496 and No. 61976219).

{\small
\bibliographystyle{ieee_fullname}
\bibliography{egbib}
}
\clearpage



\section*{Supplementary material}

# Appendix A for EigenGAN

In Sec. 1-8, we derive the analytical result of maximum likelihood estimation (MLE) for the linear case of the proposed EigenGAN. Sec. 8 discusses the MLE result and the relation among the linear EigenGAN, the Principal Component Analysis (PCA) [1], and the Probabilistic PCA [2].

## 1. The Likelihood

The linear EigenGAN relates a $d$-dimension observation vector $\mathbf{x}$ to a corresponding $q$-dimension ($q \leq d$) latent variables $\mathbf{z}$ by an affine transform $\mathbf{UL}$ and a translation $\boldsymbol{\mu}$, which is formulated as

$$\mathbf{x} = \mathbf{ULz} + \boldsymbol{\mu} + \sigma\boldsymbol{\epsilon}, \tag{1}$$

with constraints:

$$\mathbf{z} \sim \mathcal{N}_q(\mathbf{0}, \mathbf{I}), \tag{2}$$
$$\boldsymbol{\epsilon} \sim \mathcal{N}_d(\mathbf{0}, \mathbf{I}), \text{ independent of } \mathbf{z}, \tag{3}$$
$$\mathbf{U}^\mathrm{T}\mathbf{U} = \mathbf{I}, \mathbf{U} \text{ is of size } d \times q, \tag{4}$$
$$\mathbf{L} \text{ is a } q \times q \text{ diagonal matrix.} \tag{5}$$

The noise vector $\boldsymbol{\epsilon}$ in Eq. (1) is introduced to compensate the missing energy (missing rank) since the rank of the latent variables is no more than the rank of the observation ($q \leq d$). According to Eq. (1)-(3), the probability density function of $\mathbf{x}$ is

$$p_\mathbf{x}(\mathbf{x}) = \frac{1}{\sqrt{(2\pi)^d |\mathbf{C}|}} \exp\left(-\frac{1}{2}(\mathbf{x} - \boldsymbol{\mu})^\mathrm{T} \mathbf{C}^{-1}(\mathbf{x} - \boldsymbol{\mu})\right), \tag{6}$$

where

$$\mathbf{C} = \mathbf{UL}^2\mathbf{U}^\mathrm{T} + \sigma^2\mathbf{I}, \tag{7}$$
$$\mathbf{C}^{-1} = \mathbf{UMU}^\mathrm{T} + \sigma^{-2}\mathbf{I}, \tag{8}$$
$$\mathbf{M} = \left(\mathbf{L}^2 + \sigma^2\mathbf{I}\right)^{-1} - \sigma^{-2}\mathbf{I}. \tag{9}$$

Then for $n$ observations $\{\mathbf{x}\}_{i=1}^{n}$, the log-likelihood is

$$\mathcal{L}_1 = -\frac{n}{2}\left\{d\log 2\pi + \log|\mathbf{C}| + \frac{1}{n}\sum_{i=1}^{n}(\mathbf{x}_i - \boldsymbol{\mu})^{\mathrm{T}}\mathbf{C}^{-1}(\mathbf{x}_i - \boldsymbol{\mu})\right\}. \tag{10}$$

According to Eq. (7), only the square value of $\mathbf{L}$ affects the probability density function, therefore we can assume the elements of $\mathbf{L}$ to be non-negative. Further, for convenience of the following analysis, without loss of generality, we organize $\mathbf{L}$ by grouping and sorting it by the value of the diagonal elements:

$$\mathbf{L} = \mathrm{diag}\left(l_1\mathbf{I}_{d_1}, l_2\mathbf{I}_{d_2}, \cdots, l_p\mathbf{I}_{d_p}\right), \tag{11}$$

where $l_1 > l_2 > \cdots > l_p \geq 0$; $\mathbf{I}_{d_j}$ denotes a $d_j \times d_j$ identity matrix, $d_j \neq 0$, and $d_1 + d_2 + \cdots + d_p = q$. According to Eq. (9) and (11), $\mathbf{M}$ also has a grouped form:

$$\mathbf{M} = \mathrm{diag}\left(\left((l_1^2 + \sigma^2)^{-1} - \sigma^{-2}\right)\mathbf{I}_{d_1}, \cdots, \left((l_p^2 + \sigma^2)^{-1} - \sigma^{-2}\right)\mathbf{I}_{d_p}\right). \tag{12}$$

And we can also define a block form of $\mathbf{U}$ accordingly:

$$\mathbf{U} = [\mathbf{U}_1, \mathbf{U}_2, \cdots, \mathbf{U}_p], \tag{13}$$

where $\mathbf{U}_i$ is of size $d \times d_i$.

## 2. Determination of $\boldsymbol{\mu}$

The partial derivative of the log-likelihood $\mathcal{L}_1$ (10) with respect to $\boldsymbol{\mu}$ is

$$\frac{\partial \mathcal{L}_1}{\partial \boldsymbol{\mu}} = \sum_{i=1}^{n}\mathbf{C}^{-1}(\mathbf{x}_i - \boldsymbol{\mu}). \tag{14}$$

Then the stationary point is

$$\boldsymbol{\mu} = \frac{1}{n}\sum_{i=1}^{n}\mathbf{x}_i = \bar{\mathbf{x}}. \tag{15}$$

Since $\mathcal{L}_1$ is a concave function of $\boldsymbol{\mu}$, the above stationary point is also the global maximum point.

## 3. Determination of U: Part (1)

Substituting Eq. (15) into the log-likelihood $\mathcal{L}_1$ (10), we obtain a new objective:
$$\mathcal{L}_2 = -\frac{n}{2}\left\{d\log 2\pi + \log|\mathbf{C}| + \text{tr}\left(\mathbf{SC}^{-1}\right)\right\}, \tag{16}$$
where
$$\mathbf{S} = \frac{1}{n}\sum_{i=1}^{n}(\mathbf{x}_i - \bar{\mathbf{x}})(\mathbf{x}_i - \bar{\mathbf{x}})^\text{T}, \tag{17}$$
i.e., the covariance matrix of the data. According to Eq. (4), the maximization of $\mathcal{L}_2$ (16) with respect to $\mathbf{U}$ is a constrained optimization as below:
$$\max_{\mathbf{U}} \quad \mathcal{L}_2$$
$$\text{subject to} \quad \mathbf{U}^\text{T}\mathbf{U} = \mathbf{I}$$
Introducing the Lagrange multiplier $\mathbf{H}$, the Lagrangian function is
$$\begin{aligned}\mathcal{L}_\mathbf{U} &= \mathcal{L}_2 + \text{tr}\left(\mathbf{H}^\text{T}\left(\mathbf{U}^\text{T}\mathbf{U} - \mathbf{I}\right)\right)\\ &= -\frac{n}{2}\left\{d\log 2\pi + \log|\mathbf{C}| + \text{tr}\left(\mathbf{SC}^{-1}\right)\right\} + \text{tr}\left(\mathbf{H}^\text{T}\left(\mathbf{U}^\text{T}\mathbf{U} - \mathbf{I}\right)\right).\end{aligned} \tag{18}$$
Then the partial derivative of $\mathcal{L}_\mathbf{U}$ with respect to $\mathbf{U}$ is
$$\frac{\partial \mathcal{L}_\mathbf{U}}{\partial \mathbf{U}} = -n\left\{\mathbf{U}\left(\left(\mathbf{L}^2 + \sigma^2\mathbf{I}\right)^{-1}\mathbf{L}^2 + \frac{\mathbf{H} + \mathbf{H}^\text{T}}{2}\right) + \text{SUM}\right\}. \tag{19}$$
At the stationary point,
$$\text{SUM} = -\mathbf{U}\left(\left(\mathbf{L}^2 + \sigma^2\mathbf{I}\right)^{-1}\mathbf{L}^2 + \frac{\mathbf{H} + \mathbf{H}^\text{T}}{2}\right). \tag{20}$$
Left multiplying the above equation by $\mathbf{U}^\text{T}$ and using $\mathbf{U}^\text{T}\mathbf{U} = \mathbf{I}$, we obtain
$$\mathbf{U}^\text{T}\text{SUM} = -\left(\mathbf{L}^2 + \sigma^2\mathbf{I}\right)^{-1}\mathbf{L}^2 - \frac{\mathbf{H} + \mathbf{H}^\text{T}}{2}. \tag{21}$$
The right-hand side of above equation is a symmetric matrix, therefore the left-hand side $\mathbf{U}^\text{T}\text{SUM}$ is also symmetric. Furthermore, since both $\mathbf{U}^\text{T}\mathbf{SU}$ and $\mathbf{M}$ are also symmetric, to satisfy the symmetry of $\mathbf{U}^\text{T}\text{SUM}$, according to the form of $\mathbf{M}$ in Eq. (12), $\mathbf{U}^\text{T}\mathbf{SU}$ must have a similar block diagonal form:
$$\begin{aligned}\mathbf{U}^\text{T}\mathbf{SU} &= \text{diag}\left(\mathbf{A}_1, \mathbf{A}_2, \cdots, \mathbf{A}_p\right) \tag{22}\\ &= \text{diag}\left(\mathbf{Q}_1^\text{T}\mathbf{\Lambda}_1\mathbf{Q}_1, \mathbf{Q}_2^\text{T}\mathbf{\Lambda}_2\mathbf{Q}_2, \cdots, \mathbf{Q}_p^\text{T}\mathbf{\Lambda}_p\mathbf{Q}_p\right), \tag{23}\end{aligned}$$

where $\mathbf{A}_j$ is a $d_j \times d_j$ symmetric matrix, and $\mathbf{Q}_j^\mathrm{T} \mathbf{\Lambda}_j \mathbf{Q}_j$ is the eigendecomposition of $\mathbf{A}_j$. Using Eq. (20), (21), and (23), we can derive

$$\mathbf{S}\mathbf{U}\mathbf{M} = \mathbf{U} \cdot \mathrm{diag}\left(\mathbf{Q}_1^\mathrm{T} \mathbf{\Lambda}_1 \mathbf{Q}_1, \mathbf{Q}_2^\mathrm{T} \mathbf{\Lambda}_2 \mathbf{Q}_2, \cdots, \mathbf{Q}_p^\mathrm{T} \mathbf{\Lambda}_p \mathbf{Q}_p\right) \cdot \mathbf{M}. \quad (24)$$

Substituting Eq. (12) and (13) into Eq. (24), we obtain

$$\left((l_j^2 + \sigma^2)^{-1} - \sigma^{-2}\right) \mathbf{S} \mathbf{U}_j \mathbf{Q}_j^\mathrm{T} = \left((l_j^2 + \sigma^2)^{-1} - \sigma^{-2}\right) \mathbf{U}_j \mathbf{Q}_j^\mathrm{T} \mathbf{\Lambda}_j \quad (25)$$

$$\implies \mathbf{S} \mathbf{U}_j \mathbf{Q}_j^\mathrm{T} = \mathbf{U}_j \mathbf{Q}_j^\mathrm{T} \mathbf{\Lambda}_j, \quad j = 1, 2, \cdots, p', \quad (26)$$

where

$$p' = \begin{cases} p, & l_p > 0, \\ p - 1, & l_p = 0. \end{cases} \quad (27)$$

Eq. (26) tells us that, the columns of $\mathbf{U}_j \mathbf{Q}_j^\mathrm{T}$ are eigenvectors of $\mathbf{S}$, and the diagonal elements of $\mathbf{\Lambda}_j$ are the corresponding eigenvalues. Further, since $\left(\mathbf{U}_j \mathbf{Q}_j^\mathrm{T}\right)^\mathrm{T} \mathbf{U}_j \mathbf{Q}_j^\mathrm{T} = \mathbf{I}$, these eigenvectors are orthonormal. Let $\mathbf{V}_j = \mathbf{U}_j \mathbf{Q}_j^\mathrm{T}$, we obtain the stationary point:

$$\mathbf{U}_j = \mathbf{V}_j \mathbf{Q}_j, \quad j = 1, 2, \cdots, p', \quad (28)$$

where the columns of $\mathbf{V}_j$ are orthonormal eigenvectors of $\mathbf{S}$ with corresponding eigenvalues as $\mathbf{\Lambda}_j = \mathrm{diag}\left(\lambda_{j_1}, \lambda_{j_2}, \cdots, \lambda_{j_{d_j}}\right)$, and $\mathbf{Q}_j$ is an arbitrary orthogonal matrix. Note that if $p' = p - 1$, i.e., $l_p = 0$, $\mathbf{U}_p$ is an arbitrary matrix.

# 4. Determination of $\mathbf{L} = \mathrm{diag}\left(l_1 \mathbf{I}_{d_1}, l_2 \mathbf{I}_{d_2}, \cdots, l_p \mathbf{I}_{d_p}\right)$

Substituting Eq. (26), Eq. (7)-(8), and Eq. (11)-(12) into $\mathcal{L}_2$ (16) and after some manipulation, a new objective is derived:

$$\mathcal{L}_3 = -\frac{n}{2} \left\{ d \log 2\pi + \sum_{j=1}^{p'} \left[ d_j \log\left(l_j^2 + \sigma^2\right) + \left(l_j^2 + \sigma^2\right)^{-1} \mathrm{tr}\left(\mathbf{\Lambda}_j\right) \right] \right.$$
$$\left. + (d - q') \log \sigma^2 + \sigma^{-2} \left( \mathrm{tr}\left(\mathbf{S}\right) - \sum_{j=1}^{p'} \mathrm{tr}\left(\mathbf{\Lambda}_j\right) \right) \right\}, \quad (29)$$

where

$$q' = \sum_{j=1}^{p'} d_j. \quad (30)$$

Then the partial derivative of $\mathcal{L}_3$ with respect to $l_j$ is

$$\frac{\partial \mathcal{L}_3}{\partial l_j} = -n \left\{ \frac{d_j l_j}{l_j^2 + \sigma^2} - \frac{l_j \text{tr}(\mathbf{\Lambda}_j)}{(l_j^2 + \sigma^2)^2} \right\}, \quad j = 1, 2, \cdots, p', \tag{31}$$

and the stationary point is

$$l_j^2 = \frac{\text{tr}(\mathbf{\Lambda}_j)}{d_j} - \sigma^2, \quad j = 1, 2, \cdots, p'. \tag{32}$$

## 5. Determination of $\sigma$

Substituting Eq. (32) into $\mathcal{L}_3$ (29), we obtain a new objective:

$$\mathcal{L}_4 = -\frac{n}{2} \left\{ d \log 2\pi + \sum_{j=1}^{p'} \left[ d_j \log \frac{\text{tr}(\mathbf{\Lambda}_j)}{d_j} + d_j \right] \right.$$
$$\left. + (d - q') \log \sigma^2 + \sigma^{-2} \left( \text{tr}(\mathbf{S}) - \sum_{j=1}^{p'} \text{tr}(\mathbf{\Lambda}_j) \right) \right\}. \tag{33}$$

Then the partial derivative of $\mathcal{L}_4$ with respect to $\sigma$ is

$$\frac{\partial \mathcal{L}_4}{\partial \sigma} = -n \left\{ \frac{d - q'}{\sigma} - \frac{1}{\sigma^3} \left( \text{tr}(\mathbf{S}) - \sum_{j=1}^{p'} \text{tr}(\mathbf{\Lambda}_j) \right) \right\}, \tag{34}$$

and the stationary point is

$$\sigma^2 = \frac{\text{tr}(\mathbf{S}) - \sum_{j=1}^{p'} \text{tr}(\mathbf{\Lambda}_j)}{d - q'}. \tag{35}$$

## 6. Determination of $\mathbf{\Lambda}_j$

Substituting Eq. (35) into $\mathcal{L}_4$ (33), we obtain a new objective:

$$\mathcal{L}_5 = -\frac{n}{2} \left\{ d \log 2\pi + \sum_{j=1}^{p'} d_j \log \frac{\text{tr}(\mathbf{\Lambda}_j)}{d_j} + (d - q') \log \frac{\text{tr}(\mathbf{S}) - \sum_{j=1}^{p'} \text{tr}(\mathbf{\Lambda}_j)}{d - q'} + d \right\}. \tag{36}$$

According to Sec. 3, the diagonal elements of $\mathbf{\Lambda}_j = \text{diag}\left(\lambda_{j1}, \lambda_{j_2}, \cdots, \lambda_{j_{d_j}}\right), j = 1, \cdots, p'$ are the eigenvalues of $\mathbf{S}$, therefore the problem here is to select the suitable eigenvalues from $\mathbf{S}$ and separate them into different $\mathbf{\Lambda}_j$s to maximize $\mathcal{L}_5$ (36). Using Jensen's inequality:

$$\log \frac{\text{tr}(\mathbf{\Lambda}_j)}{d_j} = \log \frac{\lambda_{j_1} + \cdots + \lambda_{j_{d_j}}}{d_j} \geq \frac{\log \lambda_{j_1} + \cdots + \log \lambda_{j_{d_j}}}{d_j}, \tag{37}$$

and the equality holds if and only if $\lambda_{j_1} = \cdots = \lambda_{j_{d_j}}$. That means, no matter how we select the eigenvalues, only grouping them by the same values can maximize $\mathcal{L}_5$ (36). Therefore the optimal grouping is

$$\begin{aligned}
\mathbf{\Lambda}_j &= \text{diag}\left(\lambda_{j1}, \lambda_{j_2}, \cdots, \lambda_{j_{d_j}}\right) \\
&= \text{diag}\left(\lambda_j, \lambda_j, \cdots, \lambda_j\right) \\
&= \lambda_j \mathbf{I}_{d_j},
\end{aligned} \tag{38}$$

where $\lambda_j$ is an eigenvalue of $\mathbf{S}$ whose algebraic multiplicity $\geq d_j$.

Now, the left problem is to select the eigenvalues $\lambda_j, j = 1, \cdots, p'$. Substituting Eq. (38) into $\mathcal{L}_5$ (36), we obtain

$$\begin{aligned}
\mathcal{L}_6 &= -\frac{n}{2}\left\{d \log 2\pi + \sum_{j=1}^{p'} d_j \log \lambda_j + (d - q') \log \frac{\text{tr}(\mathbf{S}) - \sum_{j=1}^{p'} \text{tr}(\mathbf{\Lambda}_j)}{d - q'} + d\right\} \\
&= -\frac{n}{2}\left\{d \log 2\pi + \text{tr}(\log \mathbf{S}) - \sum_{i=q'+1}^{d} \log \gamma_i + (d - q') \log \frac{\sum_{i=q'+1}^{d} \gamma_i}{d - q'} + d\right\} \\
&= -\frac{n}{2}\left\{d \log 2\pi + \text{tr}(\log \mathbf{S}) - (d - q')\left(\frac{\sum_{i=q'+1}^{d} \log \gamma_i}{d - q'} - \log \frac{\sum_{i=q'+1}^{d} \gamma_i}{d - q'}\right) + d\right\}, (39)
\end{aligned}$$

where $q' = d_1 + d_2 + \cdots + d_{p'}$ and $\gamma_i, i = q' + 1, \cdots, d$ are the rest eigenvalues not

been selected. Maximizing $\mathcal{L}_6$ (39) requires maximizing

$$\mathcal{F} = \frac{\sum_{i=q'+1}^{d} \log \gamma_i}{d - q'} - \log \frac{\sum_{i=q'+1}^{d} \gamma_i}{d - q'}, \tag{40}$$

which only requires $\gamma_i, i = q' + 1, \cdots, d$ to be adjacent in ordered eigenvalues. However according to Eq. (32), we need $\lambda_j > \sigma^2, j = 1, \cdots, p'$, and then from Eq. (35), the only choice to maximize $\mathcal{F}$ is to let $\gamma_i, i = q' + 1, \cdots, d$ be the $d - q'$ smallest eigenvalues. Meanwhile, lager $q'$ leads to larger $\mathcal{F}$, therefore,

$$p' = p \tag{41}$$

$$q' = q = \sum_{j=1}^{p} d_j \tag{42}$$

## 7. Determination of U: Part (2)

According to Eq. (28) and Eq. (38), the columns of $\mathbf{V}_j$ are orthonormal eigenvectors of $\mathbf{S}$ corresponding to a same eigenvalue $\lambda_j$. Since $\mathbf{Q}_j$ is an arbitrary orthogonal matrix, the column of $\mathbf{U}_j = \mathbf{V}_j \mathbf{Q}_j$ are still orthonormal eigenvectors corresponding to the eigenvalue $\lambda_j$.

## 8. Summary and Discussion

Summarizing the above analysis (Eq. (15), (32), (35), and Sec. 7), the global maximum of the likelihood with respect to the model parameters is

$$\boldsymbol{\mu} = \frac{1}{n} \sum_{i=1}^{n} \mathbf{x}_i, \tag{43}$$

$$\sigma^2 = \frac{\text{tr}(\mathbf{S}) - \text{tr}(\boldsymbol{\Lambda})}{d - q}, \tag{44}$$

$$\mathbf{L}^2 = \boldsymbol{\Lambda} - \sigma^2 \mathbf{I}, \tag{45}$$

$$\mathbf{U} = [\mathbf{u}_1, \cdots, \mathbf{u}_q], \tag{46}$$

where the elements of the diagonal matrix $\boldsymbol{\Lambda}$ is the $q$ largest eigenvalues of the data covariance $\mathbf{S}$, and $\mathbf{u}_1, \cdots, \mathbf{u}_q$ are the principal $q$ eigenvectors corresponding to $\boldsymbol{\Lambda}$. As

can be seen, under maximum likelihood estimation, the basis vectors $\mathbf{U}$ of our linear model are exactly the same as that learned by PCA [1]. Moreover, diagonal elements of $\mathbf{L}$ represent the "importance" or "energy" of the corresponding basis vectors, and from Eq. (45), when $\sigma \to 0$, the elements of $\mathbf{L}^2$ approach the $q$ largest eigenvalues. Besides, as shown in Eq. (44), energy ($\sigma^2$) of the noise is the average of the discard eigenvalues, which exactly compensates the energy missed by the subspace model.

Our model can be viewed as constrained case of Probabilistic PCA (PPCA) [2]:

$$\mathbf{x} = \mathbf{W}\mathbf{z} + \boldsymbol{\mu} + \sigma\boldsymbol{\epsilon}, \tag{47}$$

whose maximum likelihood estimation is

$$\mathbf{W} = \mathbf{V}\left(\boldsymbol{\Lambda} - \sigma^2\mathbf{I}\right)^{\frac{1}{2}}\mathbf{Q}, \tag{48}$$

where the columns of $\mathbf{V}$ are the principal eigenvectors of the data covariance, $\boldsymbol{\Lambda}$ is a diagonal matrix whose elements are the corresponding eigenvalues, $\mathbf{Q}$ is an arbitrary orthogonal matrix. Therefore, MLE result of PPCA is nondeterministic due to the arbitrary $\mathbf{Q}$. Although $\mathbf{W}$ contains information of the principal eigenvectors, the columns of $\mathbf{W}$ itself do not show explicit property of the orthogonality. Our model (1) restricts $\mathbf{W}$ of PPCA (47) to the special form of $\mathbf{UL}$ where $\mathbf{U}$ has orthonormal columns and $\mathbf{L}$ is diagonal matrix. In consequence, MLE result of our model is deterministic (Eq. (43)-(46)). Moreover, our model can build a linear subspace with the principal eigenvectors as the basis vectors explicitly, which is very suitable for extension to the nonlinear case to learn layer-wise interpretable dimensions, as introduced in the main text.

# Appendix B for EigenGAN

## 1. All Interpretable Dimensions of Each Layer

Fig. 1 to Fig. 5 show all interpretable dimensions of each generator layer learned by the proposed EigenGAN on CelebA dataset [3], Fig. 6 to Fig. 9 show all dimensions learned on Anime dataset [1], and Fig. 10 to Fig. 14 show all dimensions learned on FFHQ dataset [2]. We traverse each dimension from $-4.5\sigma$ to $4.5\sigma$ and omit the dimensions with almost no change. "None" in these figures means the corresponding part of the dimension is uninterpretable or difficult to be assigned an attribute name. The smaller the index, the deeper the layer.

## 2. Effect of the Importance Matrix $\mathbf{L}_i$

In Sec. 3.1 of the main text, we introduce the importance matrix $\mathbf{L}_i = \text{diag}(l_{i1}, \ldots, l_{iq})$ with $l_{ij}$ deciding the importance or energy of the basis vector $\mathbf{u}_{ij}$. In Fig. 15, we compare dimensions with different importance values. As can be seen, dimensions with large importance value control large variations while dimensions with small importance value control slight variations. It cannot be judged whether the larger one of two large enough importance values controls larger variation, because it is hard to quantify the semantic variations. However, it can be sure that dimensions with small importance value can only control slight changes or even no change. Therefore, to some extent, the importance matrix can be used to select the learned semantic dimensions by discarding dimensions with small importance value. Note that the values of $\mathbf{L}_i$ across different layers ($i = 1, 2, \cdots, 6$) cannot be directly compared since they belong to different subspaces.

## 3. Network Architectures

The architectures of the generator and the discriminator of EigenGAN are shown in Fig. 16.

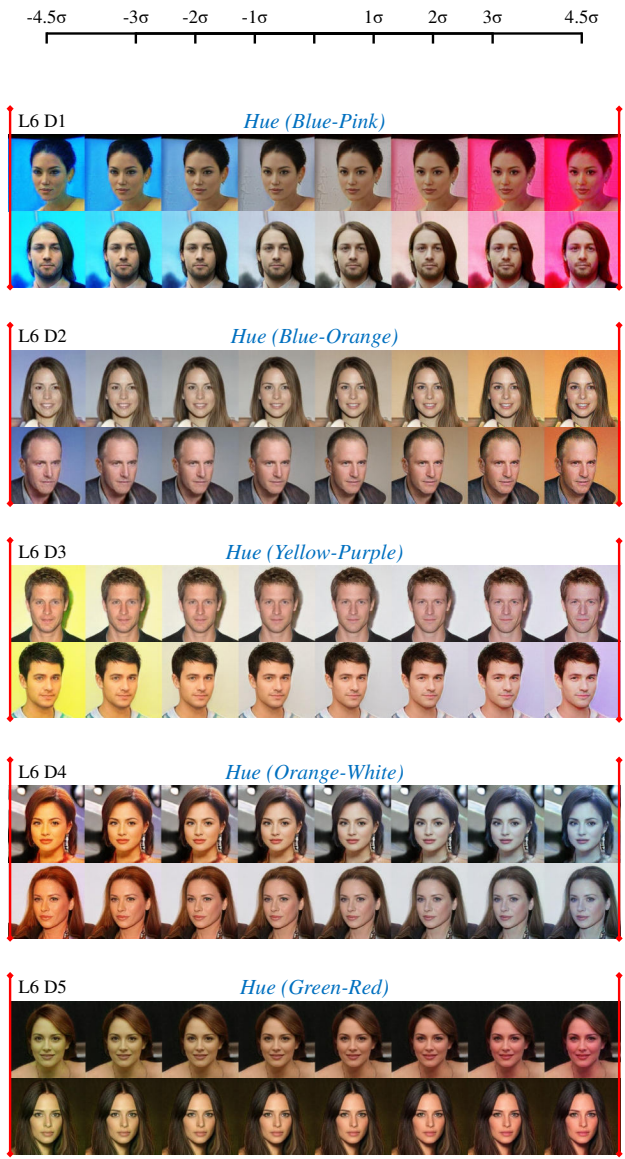

Figure 1. Interpretable dimensions of layer 6 (the shallowest) for CelebA dataset [3].

## L5 (Layer 5)

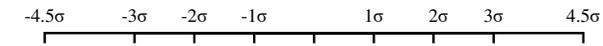

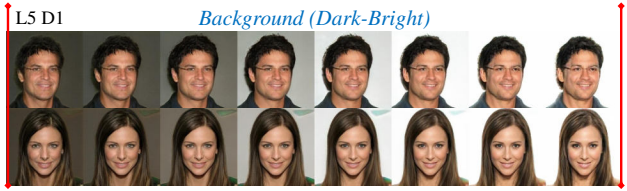

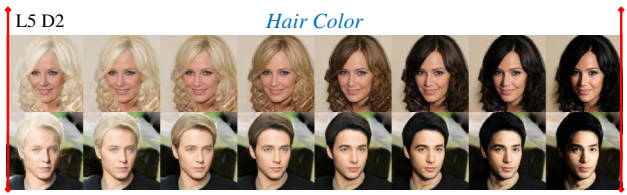

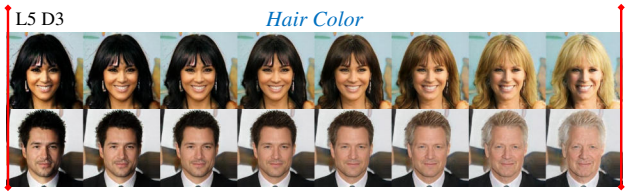

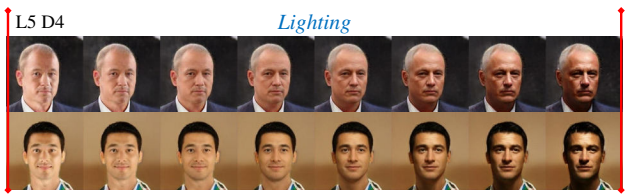

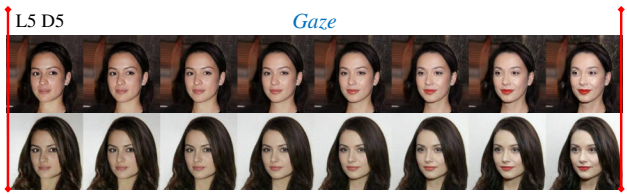

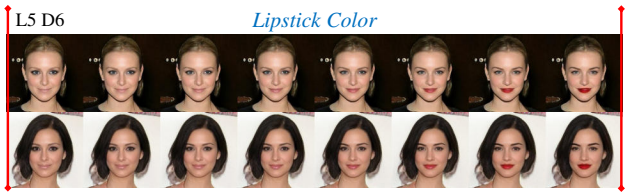

## L4 (Layer 4)

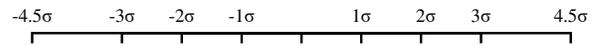

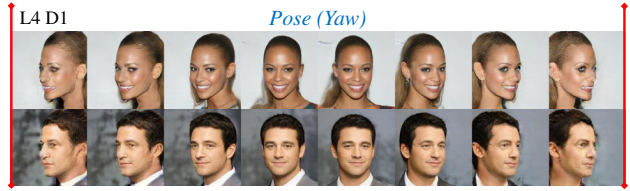

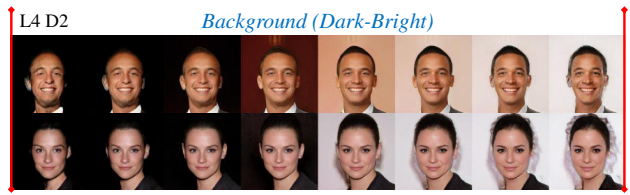

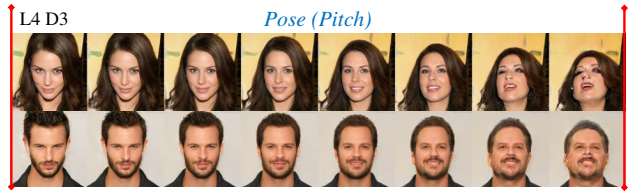

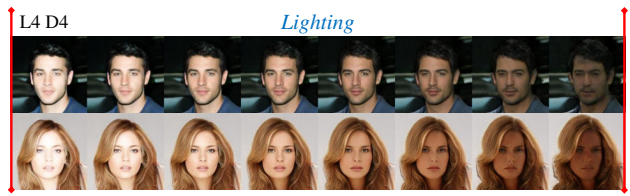

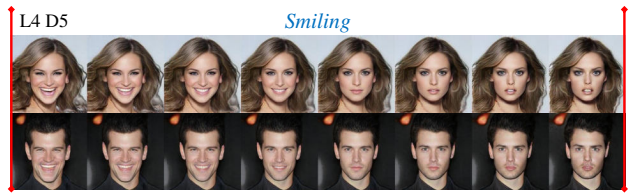

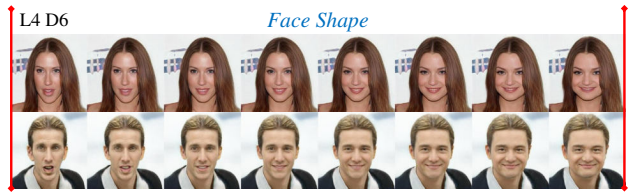

Figure 2. Interpretable dimensions of layer 5 (left) and layer 4 (right) for CelebA dataset [3].

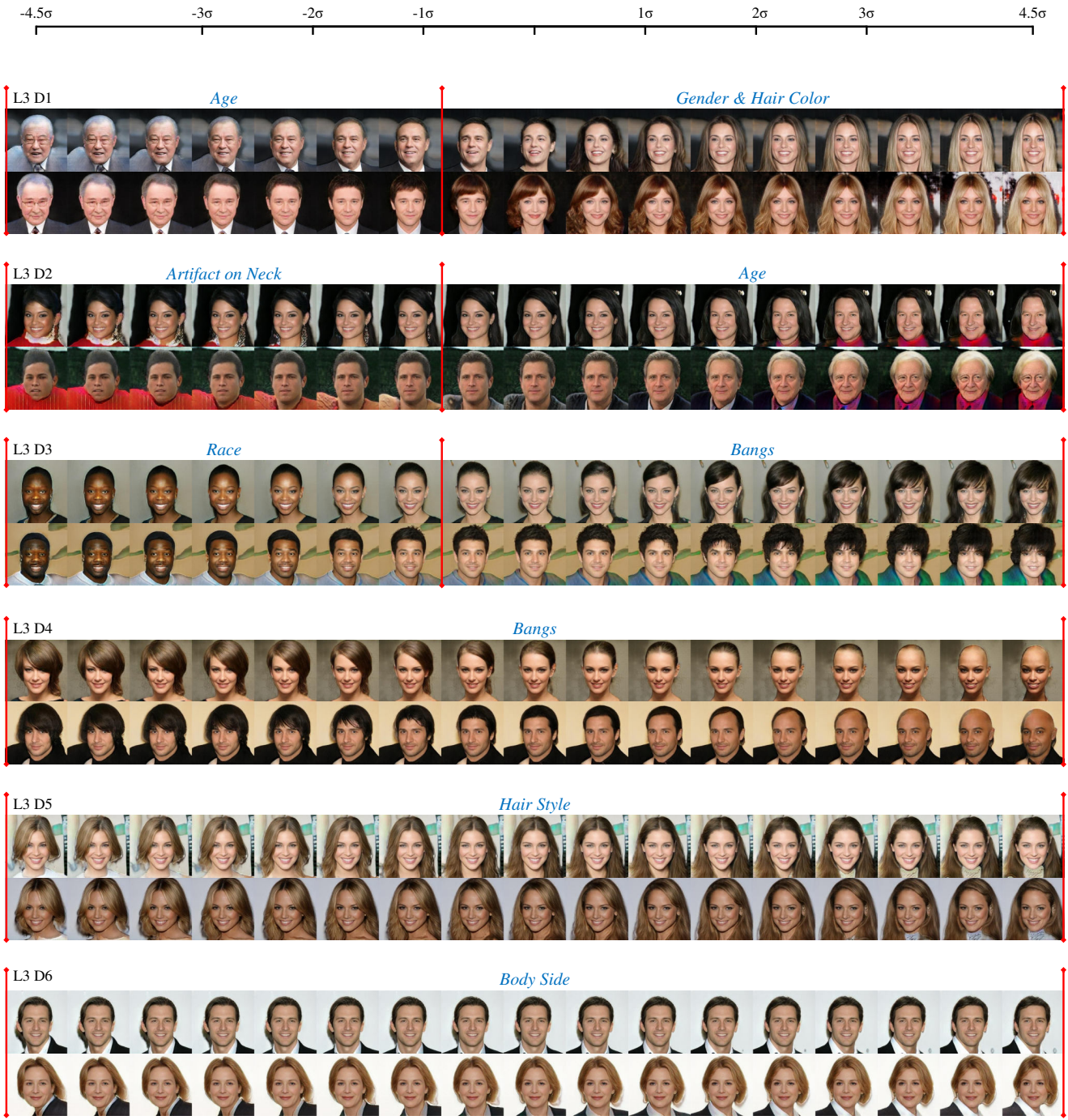

Figure 3. Interpretable dimensions of layer 3 for CelebA dataset [3].

# L2 (Layer 2)

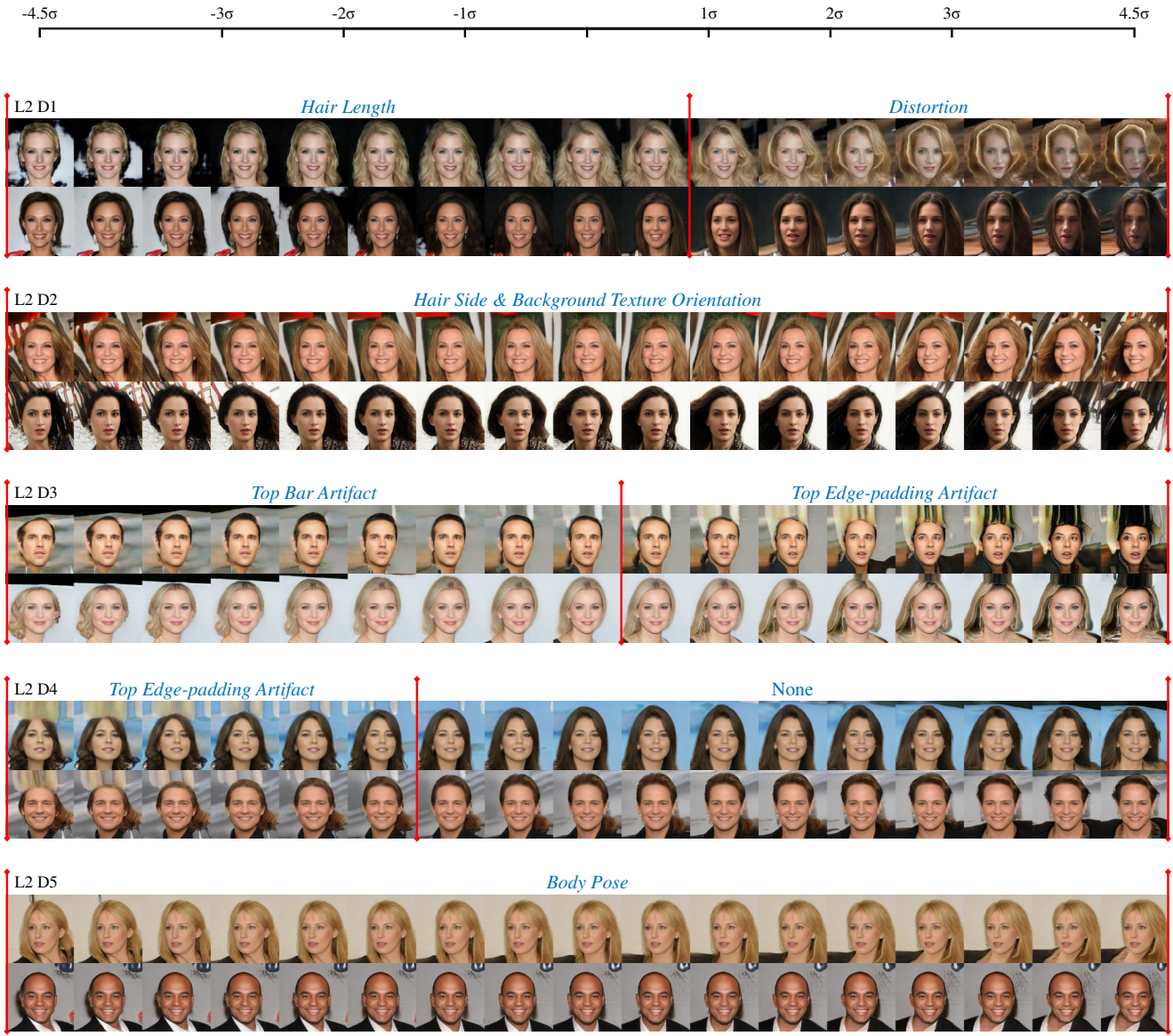

Figure 4. Interpretable dimensions of layer 2 for CelebA dataset [3]. "None" means uninterpretable.

# L1 (Layer 1, deepest)

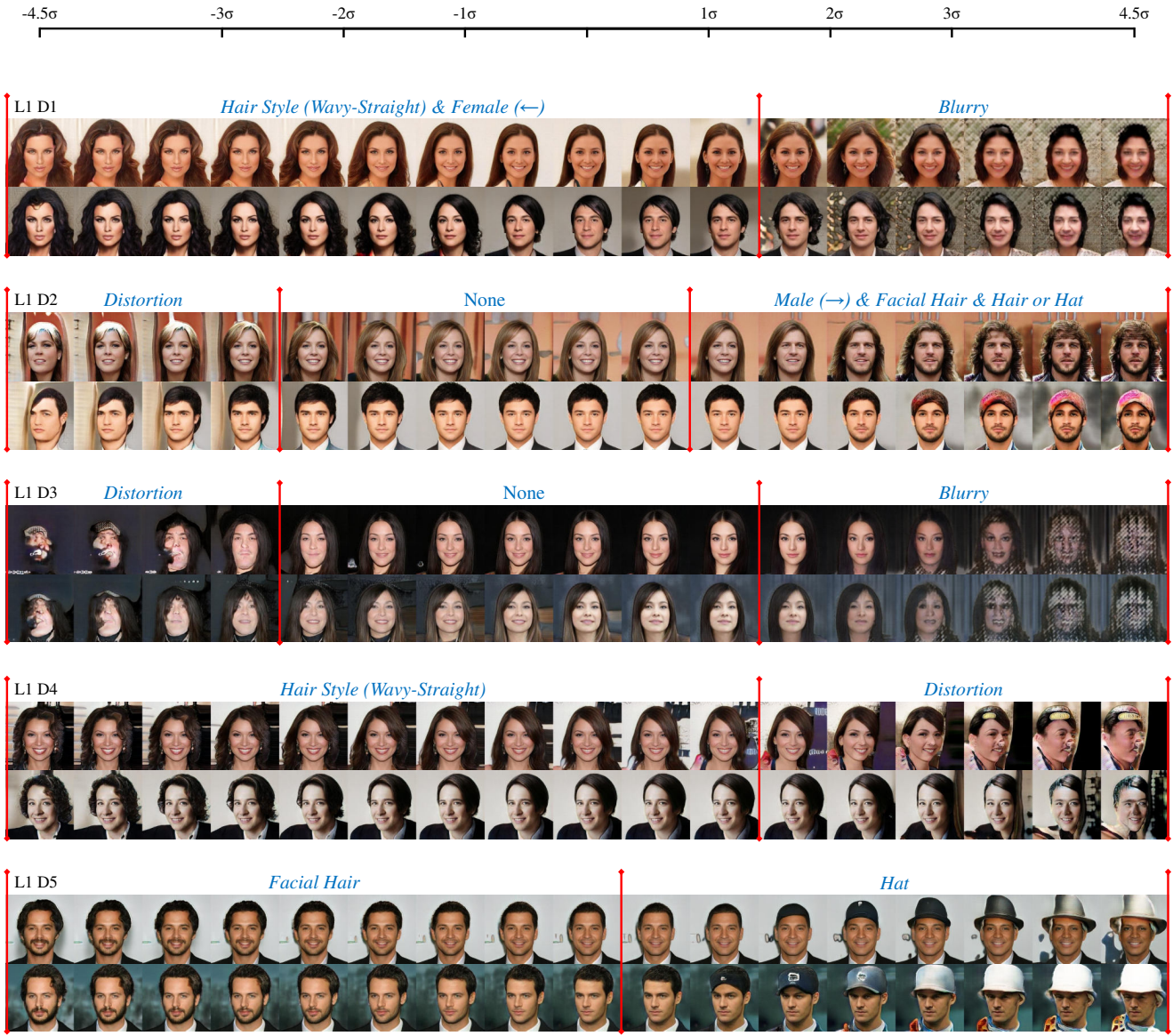

Figure 5. Interpretable dimensions of layer 1 (the deepest) for CelebA dataset [3]. "None" means uninterpretable.

## L6 (Layer 6, shallowest)

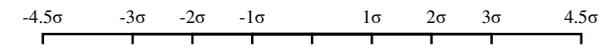

| -4.5σ | -3σ | -2σ | -1σ | | 1σ | 2σ | 3σ | 4.5σ |

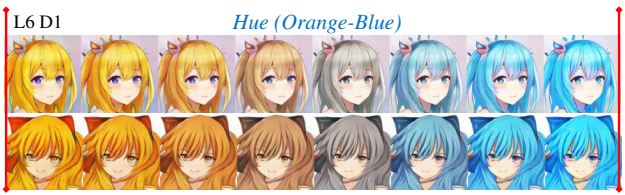

L6 D1 — *Hue (Orange-Blue)*

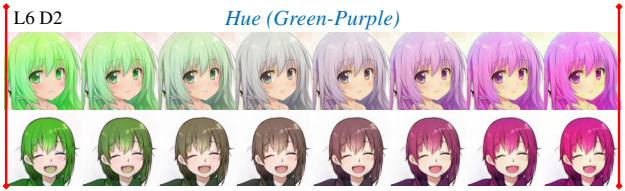

L6 D2 — *Hue (Green-Purple)*

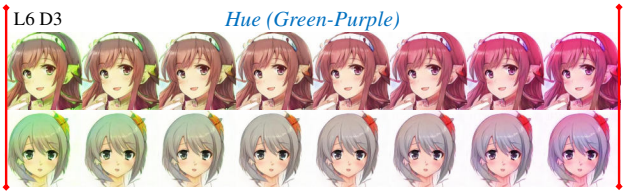

L6 D3 — *Hue (Green-Purple)*

## L5 (Layer 5)

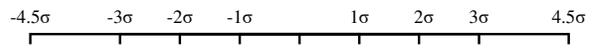

| -4.5σ | -3σ | -2σ | -1σ | | 1σ | 2σ | 3σ | 4.5σ |

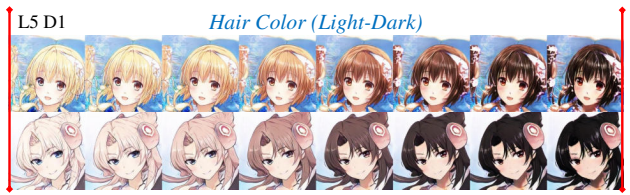

L5 D1 — *Hair Color (Light-Dark)*

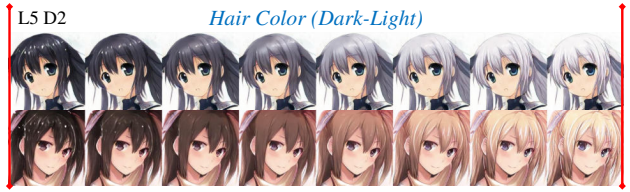

L5 D2 — *Hair Color (Dark-Light)*

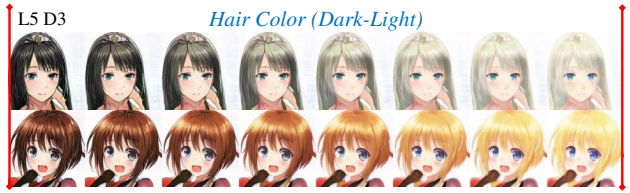

L5 D3 — *Hair Color (Dark-Light)*

Figure 6. Interpretable dimensions of layer 6 (left, the shallowest) and layer 5 (right) for Anime dataset [1].

7

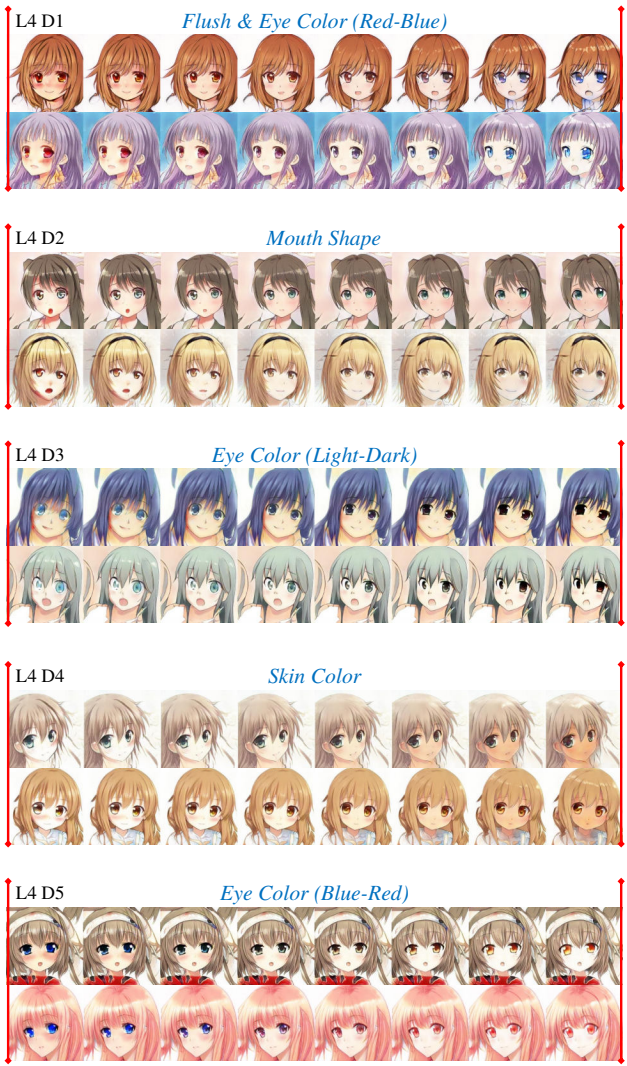
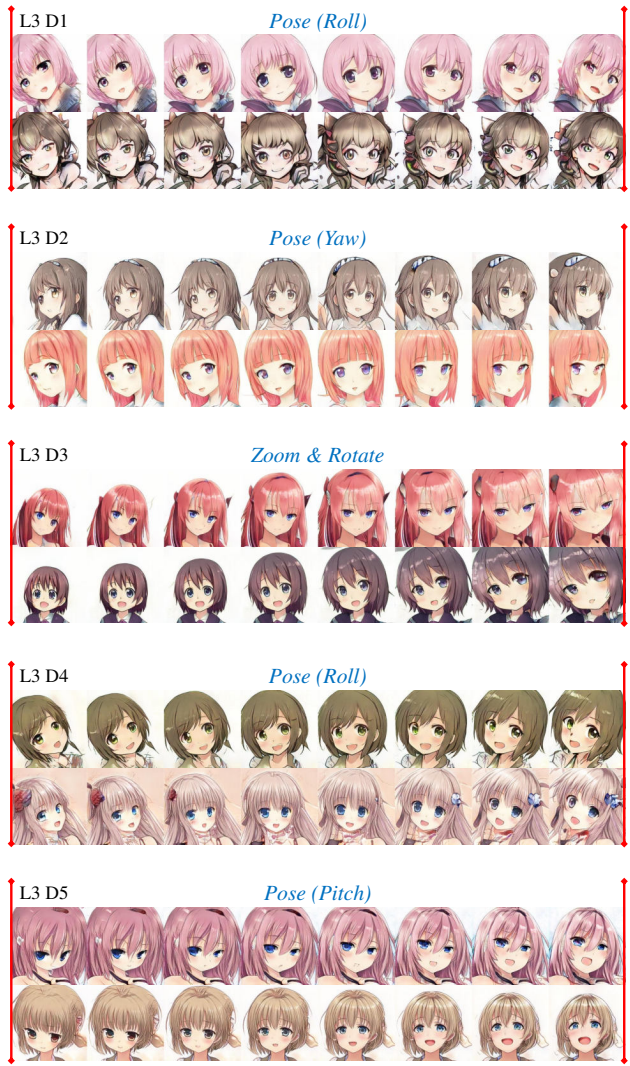

Figure 7. Interpretable dimensions of layer 4 (left) and layer 3 (right) for Anime dataset [1].

8

# L2 (Layer 2)

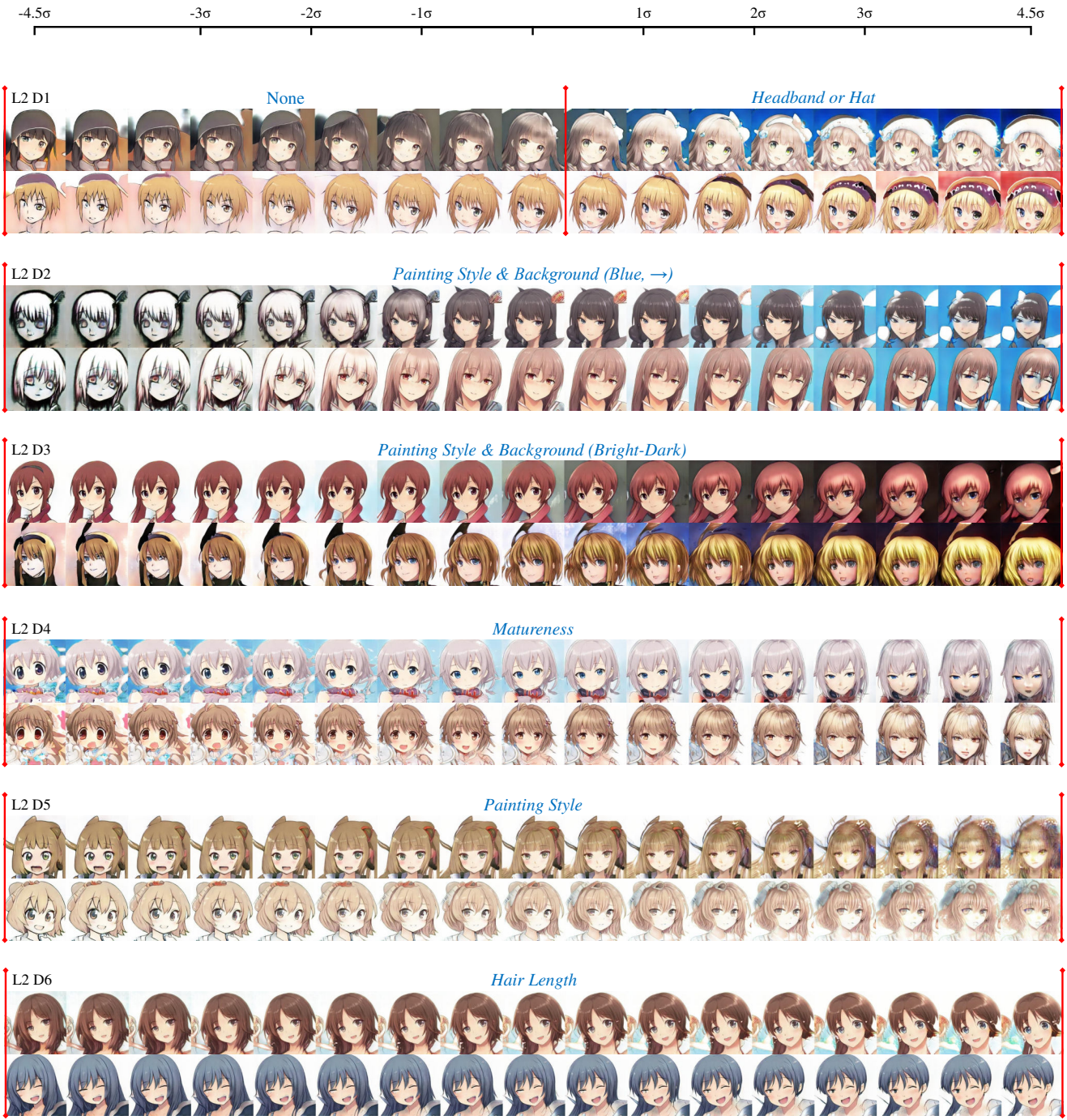

Figure 8. Interpretable dimensions of layer 2 for Anime dataset [1]. "None" means uninterpretable.

# L1 (Layer 1, deepest)

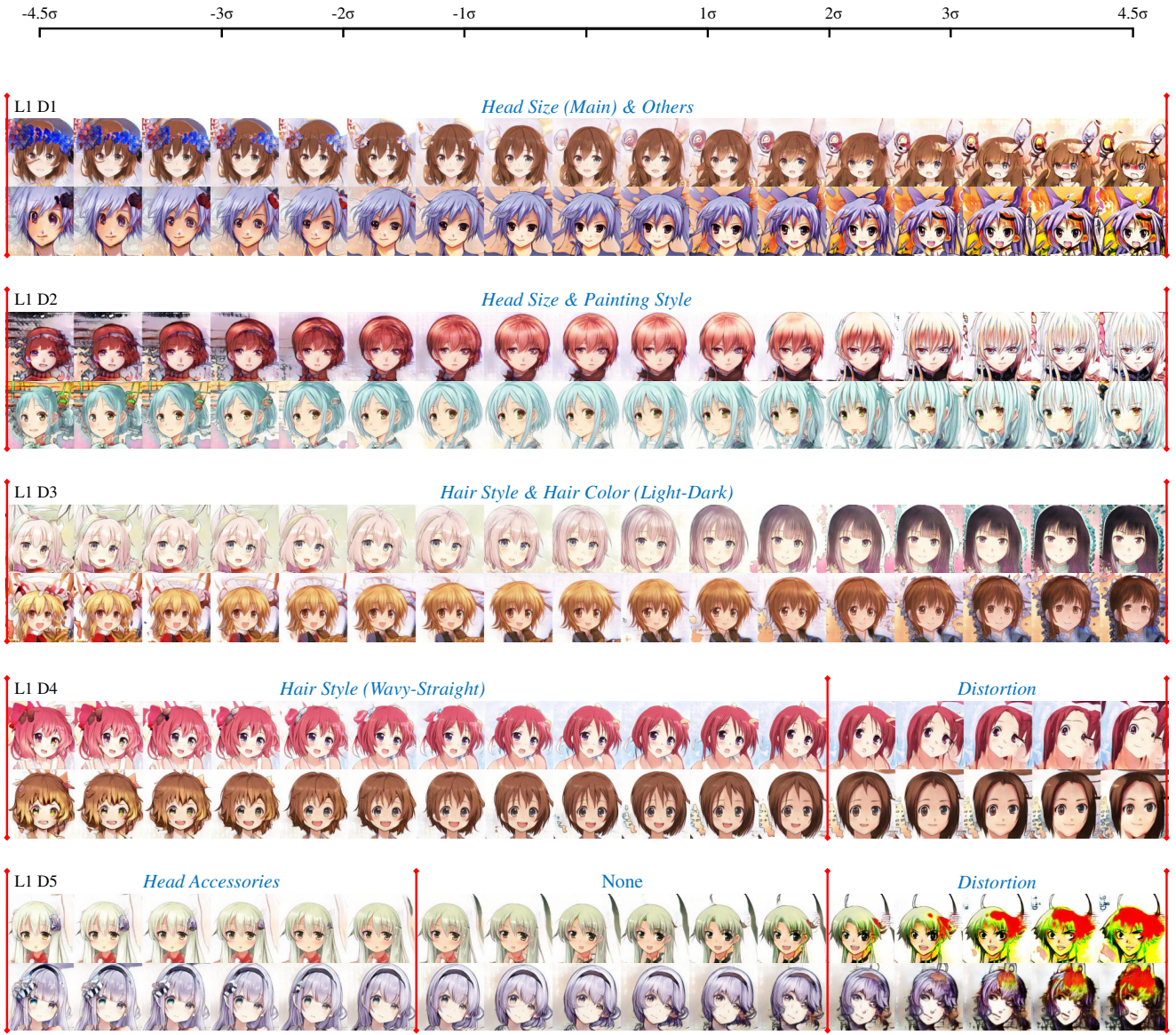

Figure 9. Interpretable dimensions of layer 1 (the deepest) for Anime dataset [1]. "None" means uninterpretable.

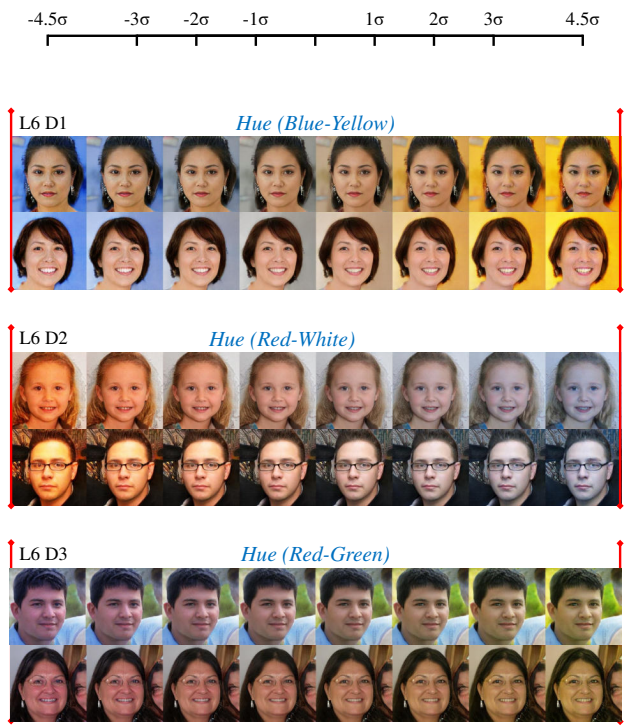

Figure 10. Interpretable dimensions of layer 6 (the shallowest) for FFHQ dataset [2].

11

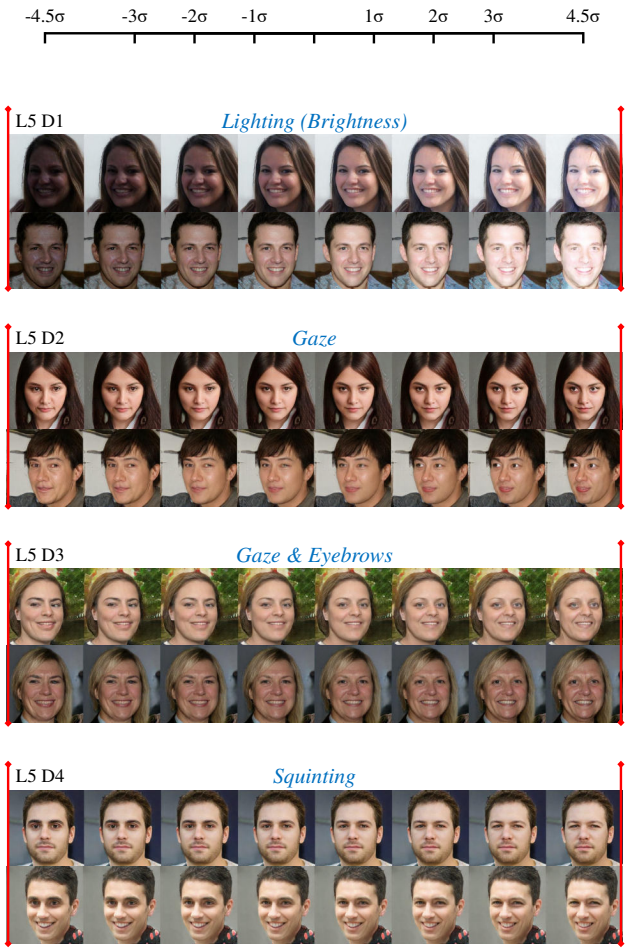
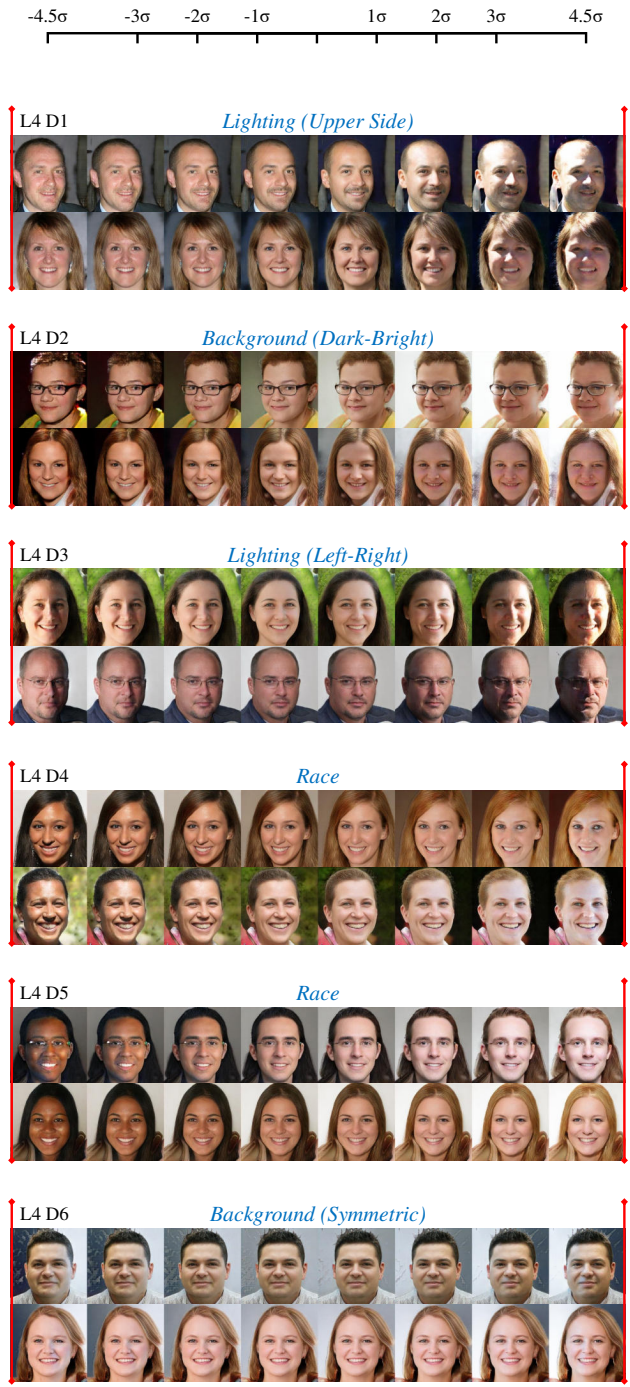

Figure 11. Interpretable dimensions of layer 5 (left) and layer 4 (right) for FFHQ dataset [2].

# L3 (Layer 3)

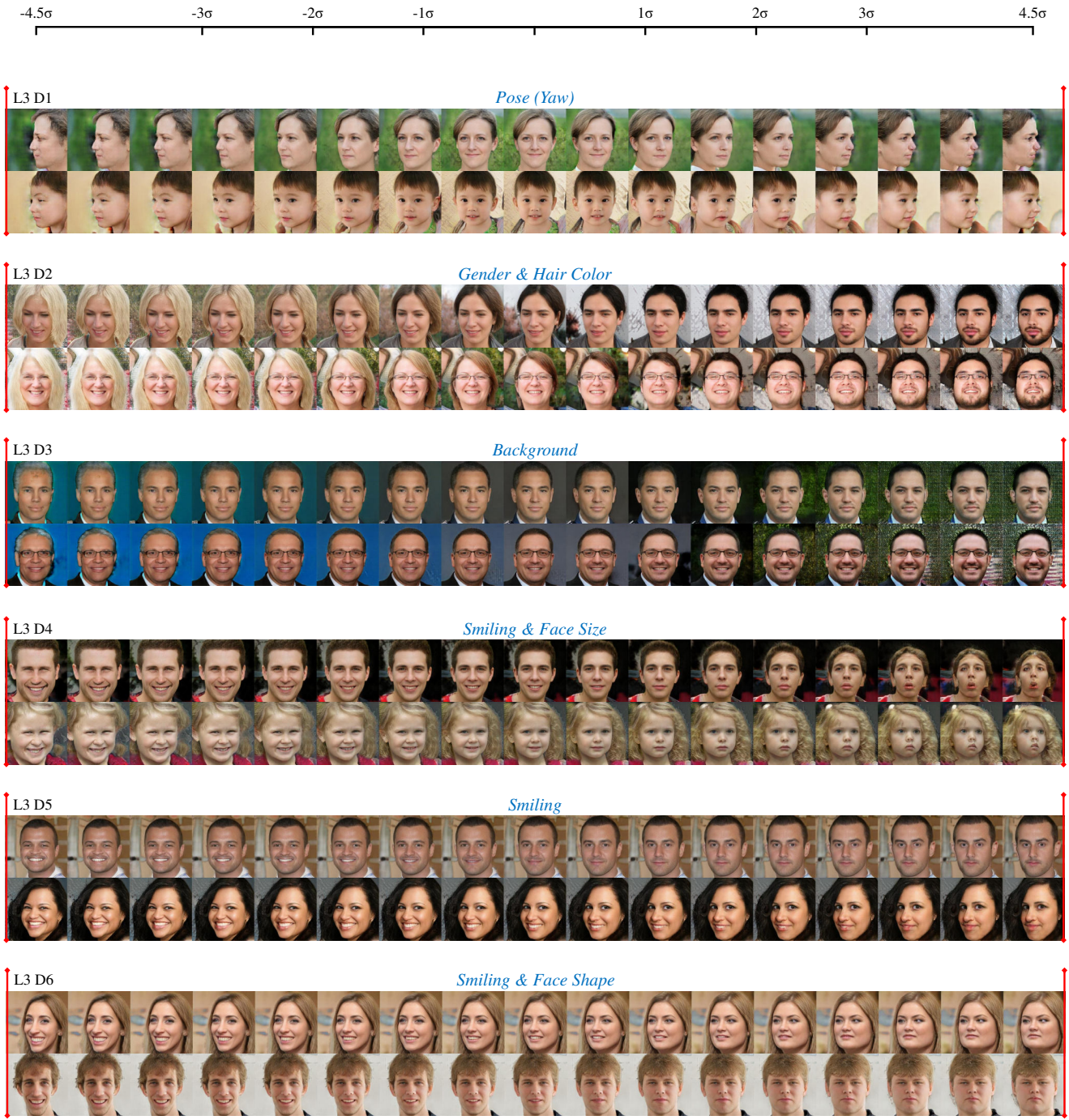

Figure 12. Interpretable dimensions of layer 3 for FFHQ dataset [2].

# L2 (Layer 2)

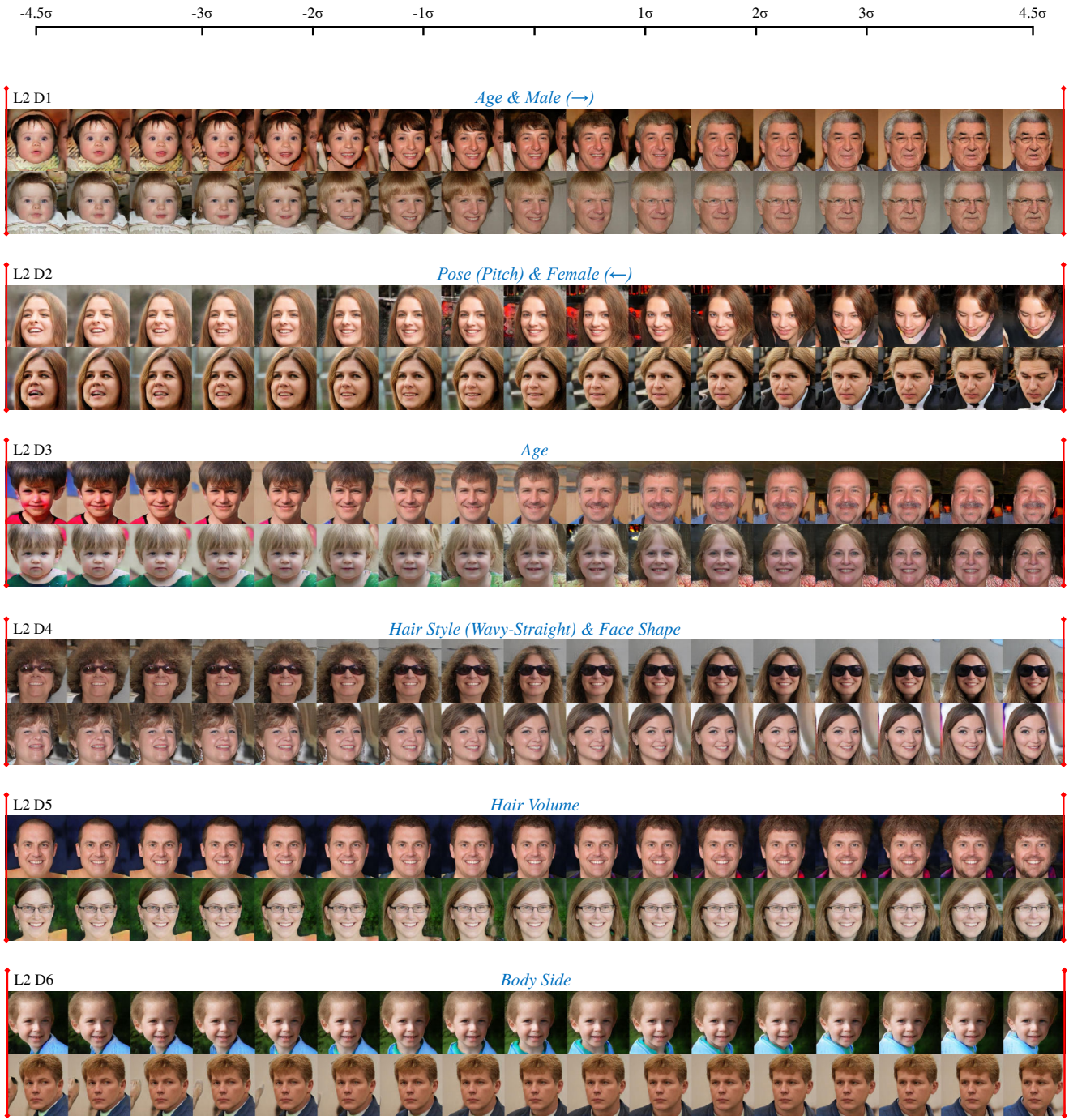

Figure 13. Interpretable dimensions of layer 2 for FFHQ dataset [2].

# L1 (Layer 1, deepest)

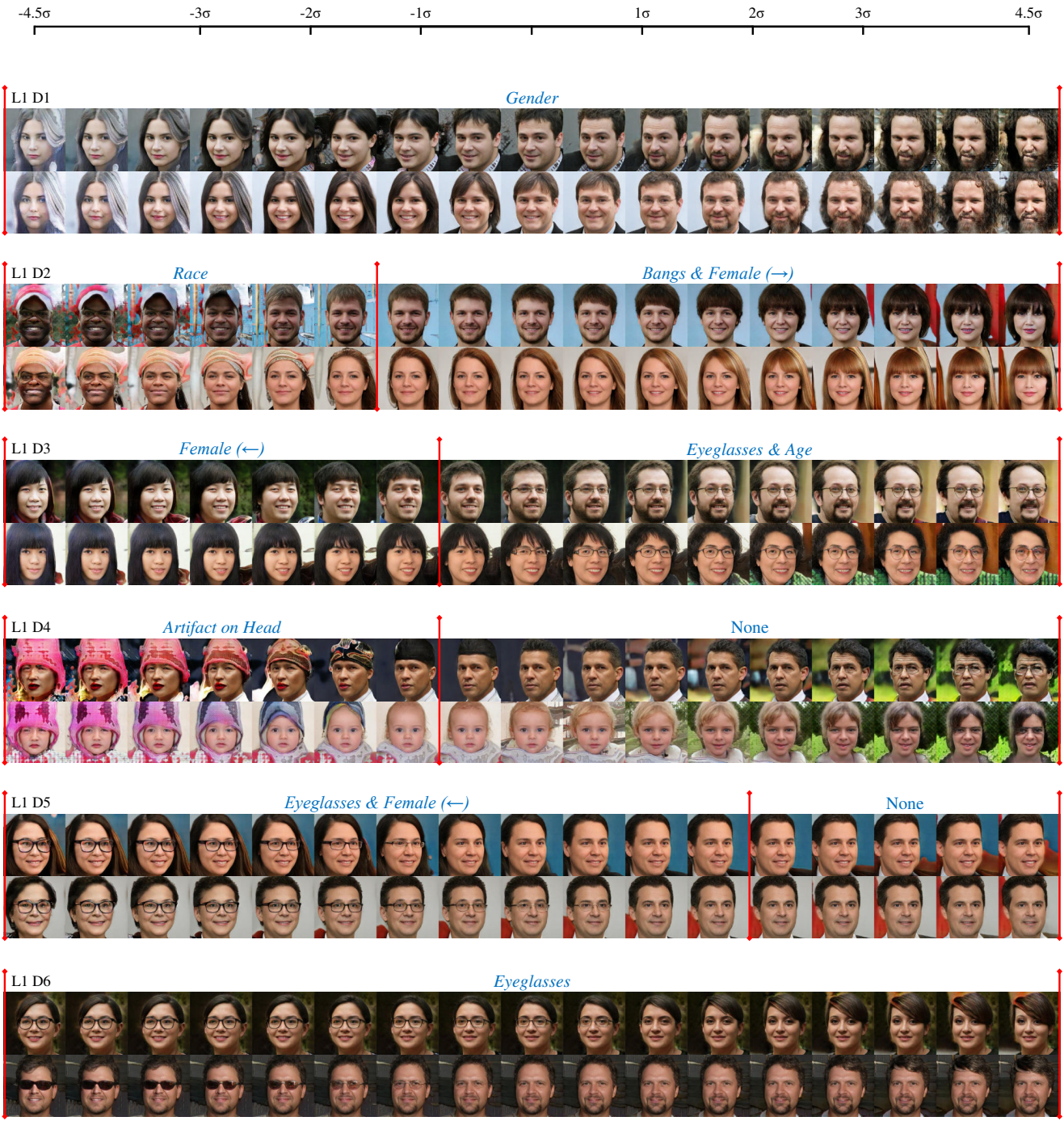

Figure 14. Interpretable dimensions of layer 1 (the deepest) for FFHQ dataset [2]. "None" means uninterpretable.

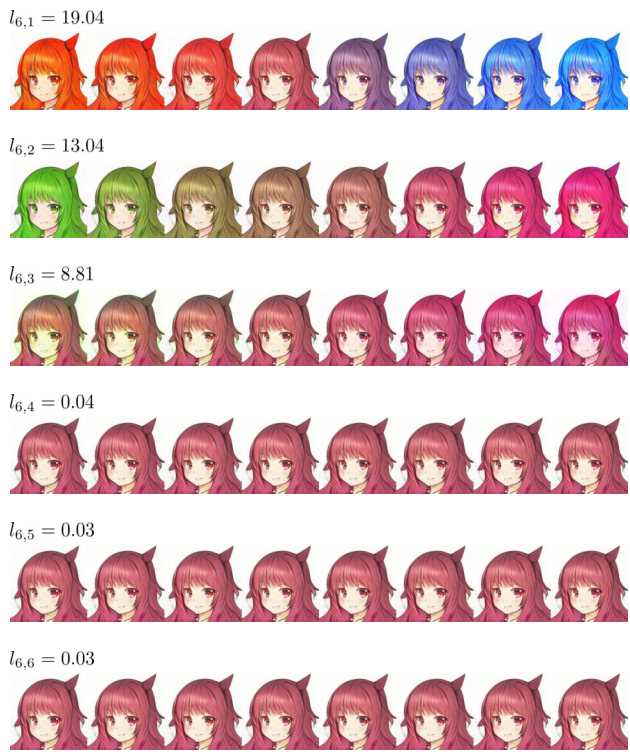
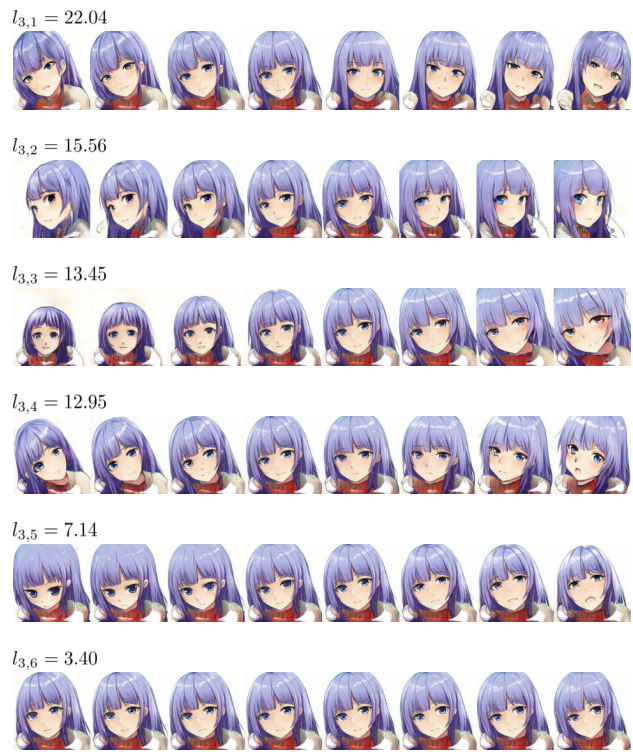

Figure 15. Effect of the importance values $\mathbf{L}_i = \text{diag}(l_{i1}, \ldots, l_{iq})$. Dimensions with large importance value control large variations while dimensions with small importance value control small variations.

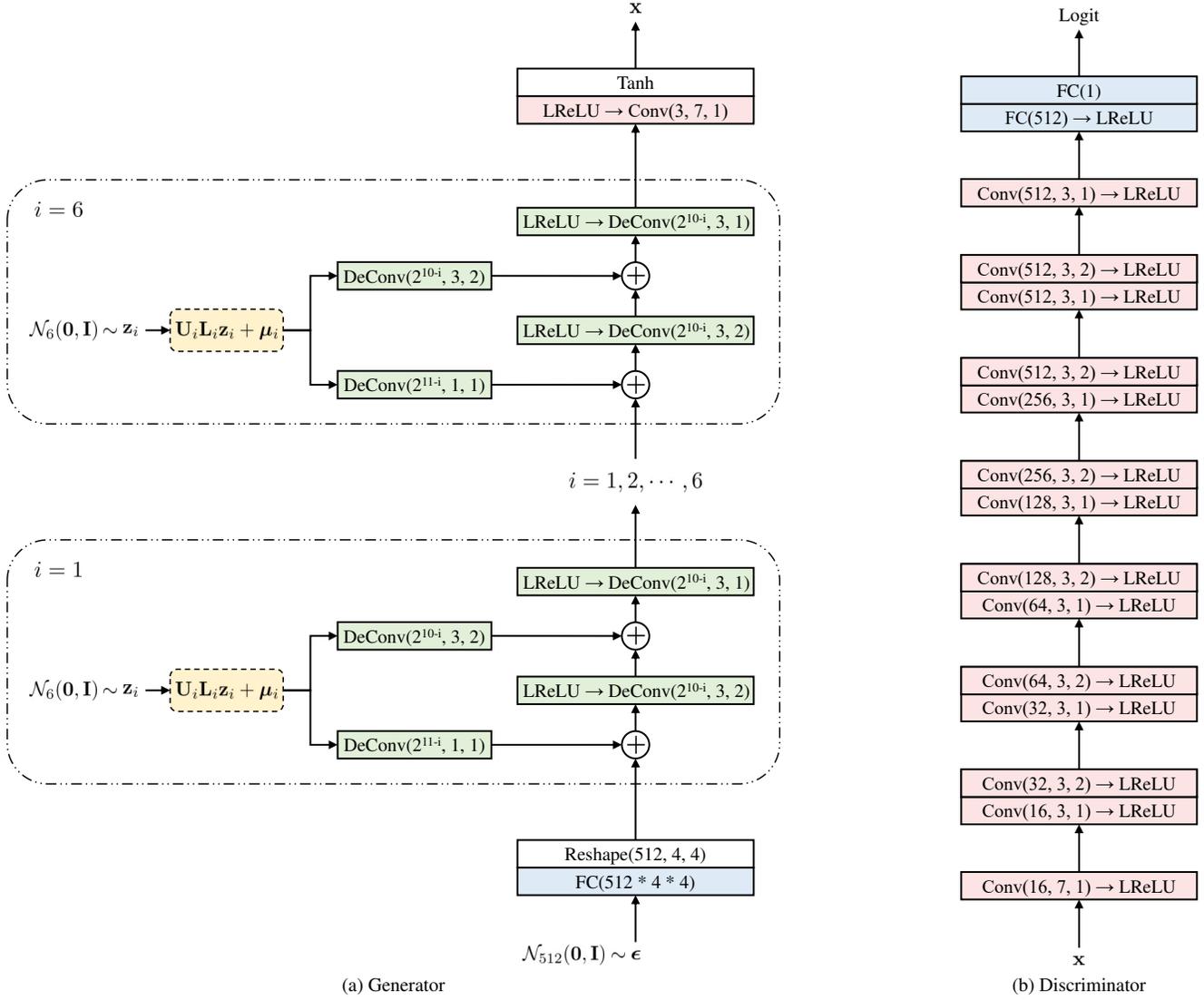

Figure 16. Network architectures of EigenGAN. Conv(d, k, s) and DeConv(d, k, s) denote convolutional layer and transposed convolutional layer with d as output dimensions, k as kernel size, and s as stride. FC(d) denotes fully connected layer with d as output dimensions. LReLU denotes Leaky ReLU [4].

17



\end{document}